\documentclass[sigconf,anonymous=false]{acmart}

\usepackage{booktabs, multirow} % For formal tables
\usepackage{amsmath}
\usepackage{array}
\usepackage{breqn}
\usepackage{graphicx}
\usepackage{algpseudocode}
\usepackage{amsthm}
\usepackage{multirow}
\usepackage{tabularx}
\usepackage{booktabs}
\usepackage{bm}
\usepackage{algorithm}
\usepackage{dsfont}
\usepackage{enumitem}
\usepackage{mathtools}
\usepackage{subfigure}
\usepackage{balance}

\usepackage{silence}
\setcopyright{iw3c2w3}

\setlist[itemize]{leftmargin=15pt}
\newcolumntype{L}[1]{>{\raggedright\let\newline\\\arraybackslash\hspace{0pt}}m{#1}}
\newcolumntype{C}[1]{>{\centering\let\newline\\\arraybackslash\hspace{0pt}}m{#1}}
\newcolumntype{R}[1]{>{\raggedleft\let\newline\\\arraybackslash\hspace{0pt}}m{#1}}
\newcolumntype{N}{@{}m{0pt}@{}}

\makeatletter
\newcommand{\multiline}[1]{%
  \begin{tabularx}{\dimexpr\linewidth-\ALG@thistlm}[t]{@{}X@{}}
    #1
  \end{tabularx}
}
\makeatother

\newcommand{\hide}[1]{}

\newcommand{\hh}[1]{{\small\color{red}{\bf hh: #1}}}
\newcommand{\qh}[1]{{\small\color{orange}{\bf QH: #1}}}
\newcommand{\kz}[1]{{\small\color{blue}{\bf KZ: #1}}}
\newcommand{\mkclean}{
	\renewcommand{\hh}[1]{}
	\renewcommand{\kz}[1]{}
    \renewcommand{\qh}[1]{}	
}

\let\oldReturn\Return
\renewcommand{\Return}{\State\oldReturn}
\theoremstyle{plain}

\newtheorem{problem}{Problem}
\graphicspath{{figures/}}

\theoremstyle{definition}

\begin{document}

\title{Few-shot Network Anomaly Detection via Cross-network Meta-learning}
\mkclean

\author{Kaize Ding$^\dagger$,~~~Qinghai Zhou$^\ddagger$,~~~Hanghang Tong$^\ddagger$,~~~Huan Liu$^\dagger$}

\thanks{The first two authors contributed equally to this work.}


\affiliation{
 \institution{$^\dagger$Arizona State University, \{kaize.ding, huan.liu\}@asu.edu \\ $^\ddagger$University of Illinois at Urbana-Champaign, \{qinghai2, htong\}@illinois.edu}
}

% \author{Kaize Ding}
% \email{kaize.ding@asu.edu}
% \affiliation{%
%   \institution{Arizona State University}
% }

% \author{Qinghai Zhou}
% \email{qinghai2@illinois.edu}
% \affiliation{%
%   \institution{University of Illinois at Urbana-Champaign}
% }
% % \authornote{Both authors contributed equally to this research.}

% \author{Hanghang Tong}
% \email{htong@illinois.edu}
% \affiliation{%
%   \institution{University of Illinois at Urbana-Champaign}
% }

% \author{Huan Liu}
% \email{huan.liu@asu.edu}
% \affiliation{%
%   \institution{Arizona State University}
% }

\renewcommand{\shortauthors}{K. Ding, et al.}

\copyrightyear{2021}
\acmYear{2021}
% \setcopyright{iw3c2w3}
\acmConference[WWW '21]{Proceedings of the Web Conference 2021}{April 19--23, 2021}{Ljubljana, Slovenia}
\acmBooktitle{Proceedings of the Web Conference 2021 (WWW '21), April 19--23, 2021, Ljubljana, Slovenia}
\acmPrice{}
\acmDOI{10.1145/3442381.3449922}
\acmISBN{978-1-4503-8312-7/21/04}

\begin{abstract}
Network anomaly detection aims to find network elements (e.g., nodes, edges, subgraphs) with significantly different behaviors from the vast majority. It has a profound impact in a variety of applications ranging from finance, healthcare to social network analysis. Due to the unbearable labeling cost, existing methods are predominately developed in an unsupervised manner. Nonetheless, the anomalies they identify may turn out to be data noises or uninteresting data instances due to the lack of prior knowledge on the anomalies of interest. Hence, it is critical to investigate and develop few-shot learning for network anomaly detection. In real-world scenarios, few labeled anomalies are also easy to be accessed on similar networks from the same domain as of the target network, while most of the existing works omit to leverage them and merely focus on a single network. Taking advantage of this potential, in this work, we tackle the problem of few-shot network anomaly detection by (1) proposing a new family of graph neural networks -- Graph Deviation Networks (GDN) that can leverage a small number of labeled anomalies for enforcing statistically significant deviations between abnormal and normal nodes on a network; (2) equipping the proposed GDN with a new cross-network meta-learning algorithm to realize few-shot network anomaly detection by transferring meta-knowledge from multiple auxiliary networks. Extensive evaluations demonstrate the efficacy of the proposed approach on few-shot or even one-shot network anomaly detection.

% with two challenges, including 
% % (1) the data heterogeneity including both the topological structure and node attributes in an attributed network, and (2) the complexity to capture information across individual source networks. 
% To address these challenges, we propose (name) for few-shot cross-network anomaly detection, where the key idea is (graph neural network based meta-learning). To be specific, (name) performs a sampling process over source networks and update the model parameters of a GNN based deviation network for anomaly detection. % The proposed algorithm bears three distinct advantages, including (1) \textit{effectiveness}, being able to remarkably improve detection accuracy on a target network;  
% We demonstrate the efficacy of the proposed approach through extensive experimental evaluations on real-world cross-network datasets.
\end{abstract}

\maketitle
\section{Introduction}\label{sec:intro}

Network-structured data, ranging from social networks~\cite{zafarani2014social} to team collaboration networks~\cite{zhou2019towards}, from citation networks~\cite{tang2008arnetminer} to molecular graphs~\cite{you2018graph}, has been widely used in modeling a myriad of real-world systems. 
Nonetheless, real-world networks are commonly contaminated with a small portion of nodes, namely, anomalies\footnote{In this paper, we primarily focus on detecting abnormal nodes.}, whose patterns significantly deviate from the vast majority of nodes~\cite{ding2019interactive,ding2020inductive,zhou2018sparc}. For instance, in a citation network that represents citation relations between papers, there are some research papers with a few spurious references (i.e., edges) which do not comply with the content of the papers~\cite{bandyopadhyay2019outlier}; In a social network that represents friendship of users, there may exist camouflaged users who randomly follow different users, rendering properties like homophily not applicable to this type of relationships~\cite{dou2020enhancing}. As the existence of even few abnormal instances could cause extremely detrimental effects, the problem of network anomaly detection has received much attention in industry and academy alike. %either industry or research community. 

% \hh{we cannot equate network anomaly = node anomaly here. In the problem definition, we can say sth like in this paper, we primarily focus on abnormal nodes.}

\begin{figure}[!t]
    \graphicspath{{figures/}}
    \centering
    \subfigure[Latent Representation Space]
    {
     \includegraphics[width=0.425\columnwidth]{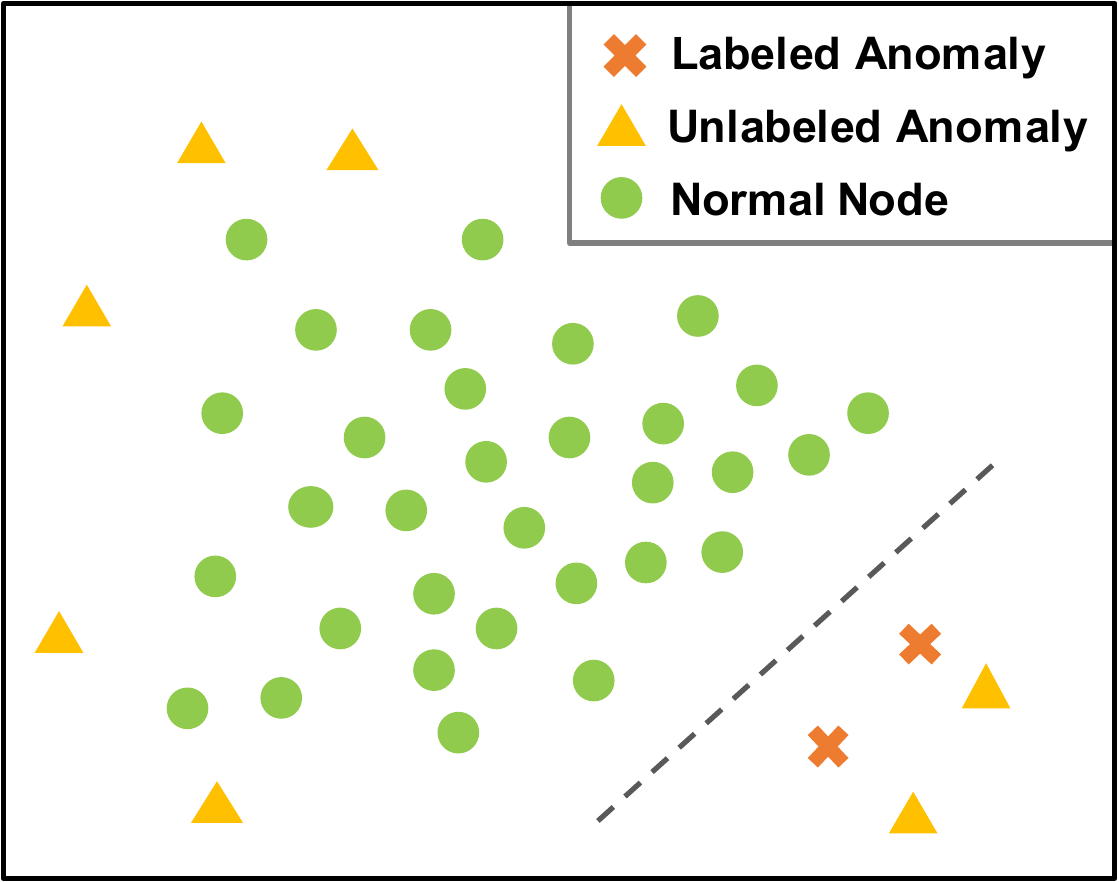}\label{fig:latent_space}
    }
    \hspace{0.1cm}
    \subfigure[Anomaly Score Space]
    {
     \includegraphics[width=0.425\columnwidth]{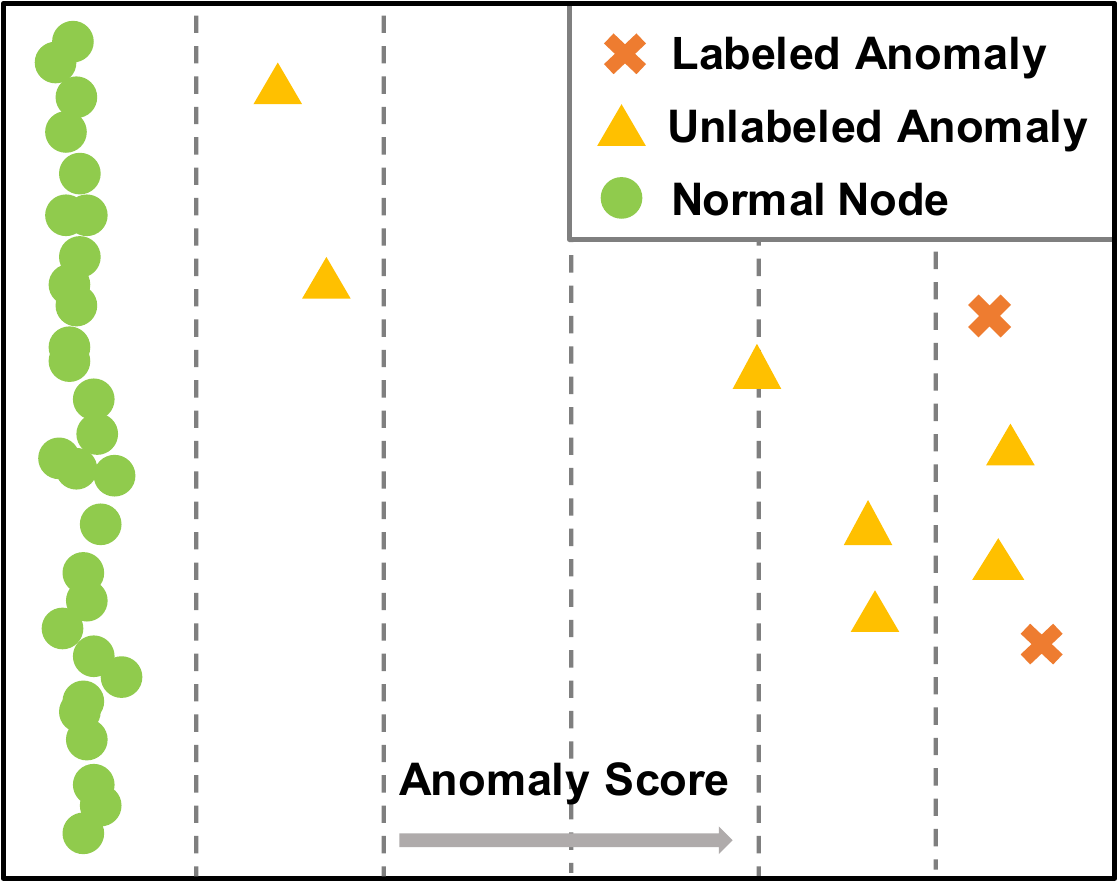}\label{fig:score_space}
    }
 \caption{Since anomalies usually have distinct patterns, (a) existing methods may easily fail to distinguish them from normal nodes in the latent representation space with only few labeled anomalies, (b) while they can be well separated in an anomaly score space  by enforcing statistically significant deviations between abnormal and normal nodes.}%
    % \label{fig:illustration}%
\end{figure}

% \hh{(1) what is 'feature representation'? (2) also need to explain what the dashed line is in (a) (3) to play the devil -- one class svm (or kernel methods in general) might work well in a}

Due to the fact that labeling anomalies is highly labor-intensive and takes specialized domain-knowledge, existing methods are predominately developed in an unsupervised manner. As a prevailing paradigm, people try to measure the abnormality of nodes with the reconstruction errors of autoencoder-based models~\cite{ding2019deep,li2019specae} or the residuals of matrix factorization-based methods~\cite{tong2011non,li2017radar,bandyopadhyay2019outlier}. However, the anomalies they identify may turn out to be data noises or uninteresting data instances due to the lack of prior knowledge on the anomalies of interest. A potential solution to this problem is to leverage limited or few labeled anomalies as the prior knowledge to learn anomaly-informed models, since it is relatively low-cost in real-world scenarios -- a small set of labeled anomalies could be either from a deployed detection system or be provided by user feedback. In the meantime, such valuable knowledge is usually scattered among other networks within the same domain of the target one, which could be further exploited for distilling supervised signal. For example, LinkedIn and Indeed have similar social networks that represent user friendship in the job-search domain; ACM and DBLP can be treated as citation networks that share similar citation relations in the computer science domain. According to previous studies~\cite{tang2019transferring,zhou2020fast,zhou2019admiring}, because of the similarity of topological structure and nodal attributes, it is feasible to transfer valuable knowledge from source network(s) to the target network so that the performance on the target one is elevated. As such, in this work we propose to investigate the novel problem of few-shot network anomaly detection under the cross-network setting.

% provides the feasibility of 
% transferring knowledge from source network(s) to target network so that the performance on the target one is elevated

Nonetheless, solving this under-explored problem remains non-trivial, mainly owing to the following reasons: \textbf{(1)} From the micro (\textit{intra-network}) view, since we only have limited knowledge of anomalies, it is hard to precisely characterize the abnormal patterns. If we directly adopt existing semi-supervised~\cite{wang2019semi} or PU~\cite{wu2019long} learning techniques, those methods often fall short in achieving satisfactory results as they might still require a relatively large percentage of positive  examples~\cite{pang2019deep}. To handle such incomplete supervision challenge~\cite{zhang2019learning} as illustrated in Figure~\ref{fig:latent_space}, instead of focusing on abnormal nodes, how to leverage labeled anomalies as few as possible to learn a high-level abstraction of normal patterns is necessary to be explored; \textbf{(2)} From the macro (\textit{inter-network}) view, though networks in the same domain might share similar characteristics in general, anomalies exist in different networks may be from very different manifolds. Previous studies on cross-network learning~\cite{wu2020unsupervised,shen2020adversarial} mostly focus on transferring the knowledge only from a single network, which may cause unstable results and the risk of negative transfer. As learning from multiple networks could provide more comprehensive knowledge about the characteristics of anomalies, a cross-network learning algorithm that is capable of adapting the knowledge is highly desirable. %required. 

% \hh{i weaken the tone a bit, as the main motivation for ssl and pu are to reduce number of labelled examples. if we say outright that 'they need many labels', some reviewers might raise eyebrowses.}

To address the aforementioned challenges, in this work we first design a new GNN architecture, namely Graph Deviation Networks (GDN), to enable network anomaly detection with limited labeled data. Specifically, given an arbitrary network, GDN first uses a GNN-backboned anomaly score learner to assign each node with an anomaly score, and then defines the mean of the anomaly scores based on a prior probability to serve as a reference score for guiding the subsequent anomaly score learning. By leveraging a deviation loss~\cite{pang2019deep}, GDN is able to enforce statistically significant deviations of the anomaly scores of anomalies from that of normal nodes in the anomaly score space (as shown in Figure~\ref{fig:score_space}). To further transfer this ability from multiple networks to the target one, we propose a cross-network meta-learning algorithm to learn a well-generalized initialization of GDN from multiple few-shot network anomaly detection tasks. The seamlessly integrated framework Meta-GDN is capable of extracting comprehensive meta-knowledge for detecting anomalies across multiple networks, which largely alleviates the limitations of transferring from a single network. Subsequently, the initialization can be easily adapted to a target network via fine-tuning with few or even one labeled anomaly, improving the anomaly detection performance on the target network to a large extent. To summarize, our main contributions is three-fold:

\begin{itemize}
    \item \textbf{\emph{Problem}}: To the best of knowledge, we are the first to investigate the novel problem of few-shot network anomaly detection. Remarkably, we propose to solve this problem by transferring the knowledge across multiple networks.

    \item \textbf{\emph{Algorithms}}: We propose a principled framework Meta-GDN, which integrates a new family of graph neural networks (i.e., GDN) and cross-network meta-learning to detect anomalies with few labeled instances.

% bridges the domain discrepancy between two attributed networks and detects both shared and unshared anomalies on the target network.
    
    \item \textbf{\emph{Evaluations}}: We perform extensive experiments to corroborate the effectiveness of our approach. The experimental results demonstrate the superior performance of Meta-GNN over the state-of-the-art methods on network anomaly detection.
\end{itemize}

 \begin{figure*}[t!]
    \graphicspath{{figures/}}
    \centering
    \includegraphics[width=1.0\textwidth]{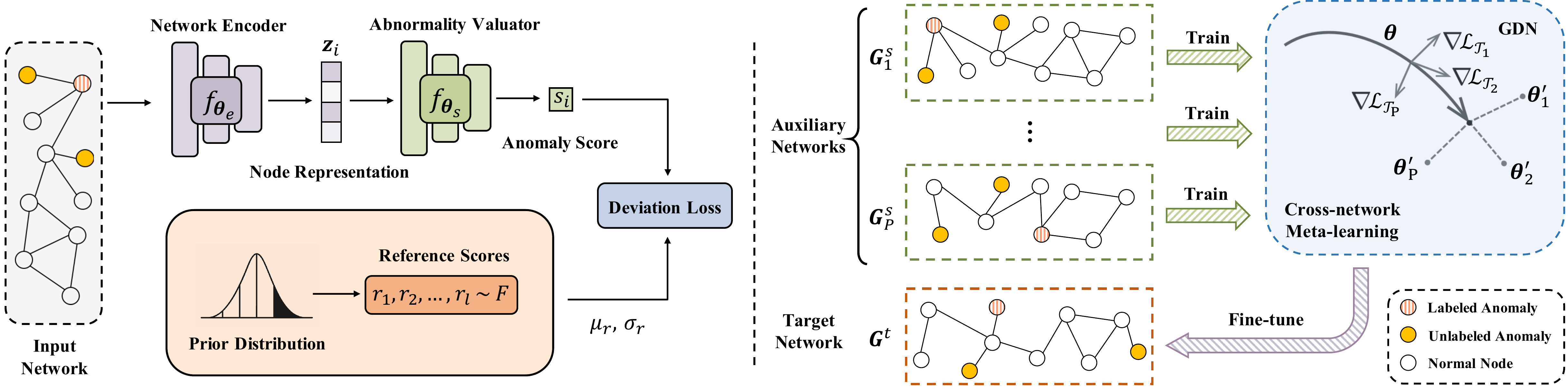}
    \caption{(Left) The model architecture of Graph Deviation Networks (GDN) for network anomaly detection with limited labeled data. (Right) The illustration of the overall framework Meta-GDN. Meta-GDN is trained across multiple auxiliary networks and can be well adapted to the target network with few-shot labeled data. Figure best viewed in color.}%
    \label{fig:framework}%
%   \vspace{-0.2cm}
\end{figure*}
\section{Related Work}\label{sec:related}
In this section, we review the related work in terms of (1) network anomaly detection; and (2) graph neural networks.
\subsection{Network Anomaly Detection}
Network anomaly detection methods have a specific focus on the network structured data. Previous research mostly study the problem of anomaly detection on plain networks. As network structure is the only available information modality in a plain network, this category of anomaly detection methods try to exploit the network structure information to spot anomalies from different perspectives~\cite{akoglu2015graph,xu2007scan}. For instance, SCAN~\cite{xu2007scan} is one of the first methods that target to find structural anomalies in networks. In recent days, attributed networks have been widely used to model a wide range of complex systems due to their superior capacity for handling data heterogeneity. In addition to the observed node-to-node interactions, attributed networks also encode a rich set of features for each node. Therefore, anomaly detection on attributed networks
has drawn increasing research attention in the community, and various methods have been proposed ~\cite{muller2013ranking,sanchez2014local}. Among them, ConOut~\cite{muller2013ranking} identifies the local context for each node and performs anomaly ranking within the local context. %FocusCO [25] focuses on community anomalies on a predefined subspace from user preferences.
More recently, researchers also propose to solve the problem of network anomaly detection using graph neural networks due to its strong modeling power. DOMINANT~\cite{ding2019deep} achieves superior performance over other shallow methods by building a deep autoencoder architecture on top of the graph convolutional networks. Semi-GNN~\cite{wang2019semi} is a semi-supervised graph neural model which adopts hierarchical attention to model the multi-view graph for fraud detection. GAS~\cite{li2019spam} is a GCN-based large-scale anti-spam
method for detecting spam advertisements. Zhao et al. propose a novel loss function to train GNNs for anomaly-detectable node representations ~\cite{zhao2020error}. Apart from the aforementioned methods, our approach focus on detecting anomalies on a target network with few labels by learning from multiple auxiliary networks.

% {\color{red}SEANO~\cite{liang2018semi} aims to learn robust embeddings that jointly preserve graph proximity, attribute affinity and label information while accounting for effects of anomalies. ONE~\cite{bandyopadhyay2019outlier} is an unsupervised attributed network embedding approach that jointly learns and minimizes the effect of anomalies in the network}. 

\subsection{Graph Neural Networks}

Graph neural networks~\cite{cao2016deep,kipf2017semi,velickovic2017graph,hamilton2017inductive} have achieved groundbreaking success in transforming the information of a graph into low-dimensional latent representations. Originally inspired by graph spectral theory, spectral-based graph convolutional networks (GCNs) have emerged and demonstrated their efficacy by designing different graph convolutional layers. Among them, The model proposed by Kipf et al.~\cite{kipf2017semi} has become the most prevailing one by using a linear filter. In addition to spectral-based graph convolution models, spatial-based graph neural networks that follow neighborhoods aggregation schemes also have been extensively investigated. Instead of training individual embeddings for each node, those methods learn a set of \textit{aggregator functions} to aggregate features from a node's local neighborhood.\hide{\qh{description}} GraphSAGE~\cite{hamilton2017inductive} learns an embedding function that can be generalized to unseen nodes, which enables inductive representation learning on network-structured data. \hide{learns a function that generates embeddings by sampling and aggregating features from a node's local neighborhood.} Similarly, Graph Attention Networks (GATs)~\cite{velickovic2017graph} proposes to learn hidden representations by introducing a self-attention strategy when aggregating neighborhood information of a node.\hide{incorporate trainable attention weights to specify fine-grained weights on neighbors.} Furthermore, Graph Isomorphism Network (GIN)~\cite{xu2018powerful} extends the idea of parameterizing universal multiset functions with neural networks\hide{ with arbitrary aggregation functions on multi-sets}, and is proven to be as theoretically powerful as the Weisfeiler-Lehman (WL) graph isomorphism test. To go beyond a single graph and transfer the knowledge across multiple ones, more recently, researchers have explored to integrate GNNs with meta-learning techniques~\cite{zugner2019adversarial,tang2019transferring,zhou2020fast}. For instance, PA-GNN~\cite{tang2019transferring} transfers the robustness from cleaned graphs to the target graph via meta-optimization. Meta-NA~\cite{zhou2020fast} is a graph alignment model that learns a unified metric space across multiple graphs, where one can easily link entities across different graphs. However, those efforts cannot be applied to our problem and we are the first to study the problem of few-shot cross-network anomaly detection.

% Transferring Robustness for Graph Neural Network Against
% Poisoning Attacks

% Graph Few-shot Learning via Knowledge Transfer

% Fast Network Alignment via Graph Meta-Learning

% FEW-SHOT LEARNING ON GRAPHS VIA SUPER-CLASSES BASED ON GRAPH SPECTRAL MEASURES

\section{Problem Definition}\label{sec:problem}
In this section, we formally define the problem of few-shot cross-network anomaly detection. Throughout the paper, we use bold uppercase letters for matrices (e.g., $\mathbf{A}$), bold lowercase letters for vectors (e.g., $\mathbf{u}$), lowercase letters for scalars (e.g., $s$) and calligraphic fonts to denote sets (e.g., $\mathcal{V}$). Notably, in this work we focus on attributed network for a more general purpose. Given an attributed network $\mathbf{G} = (\mathcal{V}, \mathcal{E}, \mathbf{X})$ where $\mathcal{V}$ is the set of nodes, i.e., $\{v_1, v_2, \dots,v_n\}$, $\mathcal{E}$ denotes the set of edges, i.e., $\{e_1, e_2, \dots, e_m\}$. The node attributes are represented by $\mathbf{X} = [\mathbf{x}_1^{\mathrm{T}}, \mathbf{x}_2^{\mathrm{T}}, \cdots, \mathbf{x}_n^{\mathrm{T}}] \in \mathbb{R}^{n\times d}$ and $\mathbf{x}_i$ is the attribute vector for node $v_i$. More concretely, we represent the attributed network as $\mathbf{G}=(\mathbf{A}, \mathbf{X})$, where  $\mathbf{A} = \{0, 1\}^{n \times n}$ is an adjacency matrix representing the network structure. Specifically, $\mathbf{A}_{i,j}=1$ indicates that there is an edge between node $v_i$ and node $v_j$; otherwise, $\mathbf{A}_{i,j}=0$.

Generally speaking, few-shot cross-network anomaly detection aims to maximally improve the detection performance on the target network through transferring very limited supervised knowledge of ground-truth anomalies from the auxiliary network(s). In addition to the target network $\mathbf{G}^t$, in this work we assume there exist $P$ auxiliary networks $ \mathcal{G}^s = \{\mathbf{G}_1^s,\mathbf{G}_2^s,\dots,\mathbf{G}_P^s \}$ sharing the same or similar domain with $\mathbf{G}^t$. For an attributed network, the set of labeled abnormal nodes is denoted as $\mathcal{V}^L$ and the set of unlabeled nodes is represented as $\mathcal{V}^U$. Note that $\mathcal{V} = \{\mathcal{V}^L, \mathcal{V}^U\}$ and in our problem $|\mathcal{V}^L| \ll|\mathcal{V}^U|$ since only few-shot labeled data is given. As network anomaly detection is commonly formulated as a ranking problem~\cite{akoglu2015graph}, we formally define the few-shot cross-network anomaly detection problem  as follows:

% In our setting, $|\mathcal{V}^L| \ll|\mathcal{V}^U|$, which means .

\begin{problem}{\textbf {Few-shot Cross-network Anomaly Detection}}\label{prob:def}
	\begin{description}
		\item[Given:] $P$ auxiliary networks, i.e., $\mathcal{G}^s = \{\mathbf{G}_1^s = (\mathbf{A}_1^s, \mathbf{X}_1^s),\mathbf{G}_2^s = (\mathbf{A}_2^s, \mathbf{X}_2^s),  \dots,\mathbf{G}_P^s =(\mathbf{A}_P^s, \mathbf{X}_P^s) \}$ and a target network $\mathbf{G}^t = (\mathbf{A}^t, \mathbf{X}^t)$, each of which contains a set of few-shot labeled anomalies (i.e., $\mathcal{V}_1^L, \mathcal{V}_2^L, \dots, \mathcal{V}_P^L$ and $\mathcal{V}^{L}_t$).
		
		\item[Goal:] to learn an anomaly detection model, which is capable of leveraging the knowledge of ground-truth anomalies from the multiple auxiliary networks, i.e., $\{\mathbf{G}_1^s,\mathbf{G}_2^s,\dots,\mathbf{G}_P^s  \}$, to detect abnormal nodes in the target network  $\mathbf{G}^t$. Ideally, anomalies that are detected should have higher ranking scores than that of the normal nodes.
		
	\end{description}
\end{problem}

\section{Proposed Approach}
In this section, we introduce the details of the proposed framework -- Meta-GDN for few-shot network anomaly detection. Specifically, Meta-GDN addresses the discussed challenges with the following two key contributions: (1) Graph Deviation Networks (GDN), a new family of graph neural networks that enable anomaly detection on an arbitrary individual network with limited labeled data; and (2) a cross-network meta-learning algorithm, which empowers GDN to transfer meta-knowledge across multiple auxiliary networks to enable few-shot anomaly detection on the target network. An overview of the  proposed Meta-GDN is provided in Figure \ref{fig:framework}.

% Two major challenges exist in the task of few-shot network anomaly detection, including (1) in many real world applications, only very limited amount of labeled anomalies is available from an individual network to train an effective machine learning model because of the exorbitant cost in obtaining sufficient data (2) anomalies from different auxiliary networks are often disparate and uncorrelated, and therefore are from extremely different manifolds, which makes it difficult for transferring learned knowledge across multiple networks.

\subsection{Graph Deviation Networks}
\label{subsec:gdn}
% \qh{c1. leveraging limited labeled data; c2. anomalies behaves differently in different network(from different manifolds), which makes it hard for cross-network learning->deviation loss}

To enable anomaly detection on an arbitrary network with few-shot labeled data, we first propose a new family of graph neural networks, called Graph Deviation Network (GDN). In essence, GDN is composed of three key building blocks, including (1) a \textit{network encoder} for learning node representations; (2) an \textit{abnormality valuator} for estimating the anomaly score for each node; and (3) a \textit{deviation loss} for optimizing the model with few-shot labeled anomalies. The details are as follows:

% a graph anomaly scoring network that encodes each node and computes anomaly scores; (2) a deviation loss function where the objective is to assign significantly greater anomaly scores to true anomalous nodes than normal nodes.

% different from normal behaviors, anomalous activities themselves are disparate and uncorrelated, which is a critical challenge to the general optimization objectives: data samples within the same category should be similar to each other

% \qh{deviation loss to distinguish between normal and abnormal data points, motivation: anomalies in different network behave differently, why using deviation  network}
% , and (3) detecting anomalies on networked data requires effectively leveraging the topological and attribute information of graphs.

% To address these challenges, we propose a novel end-to-end architecture, Graph Deviation Network, for anomaly detection on a single network, which consists of two essential component: (1) Simple Graph Convolution (SGC) that obtains node representation~\cite{wu2019simplifying}, and (2) a deviation network that learns an anomaly scoring function, detailed as follows\kz{You'd better highlight the two }.
% \qh{1. graph anomaly scoring for node representation and computing anomaly scores; 2. deviation loss}

\smallskip
\noindent\textbf{Network Encoder.} In order to learn expressive nodes representations from an input network, we first build the \textit{network encoder} module. Specifically, it is built with multiple GNN layers that encode each node to a low-dimensional latent representation. In general, GNNs follow the neighborhood message-passing mechanism, and compute the node representations by aggregating features from local neighborhoods in an iterative manner. Formally, a generic GNN layer computes the node representations using two key functions:

\begin{equation}
\begin{aligned}
    \mathbf{h}_{\mathcal{N}_i}^l &= \textsc{Aggregate}^l\Big(\{   \mathbf{h}_j^{l-1} \arrowvert \forall j \in \mathcal{N}_i \cup v_i \}\Big),\\
    \mathbf{h}_i^l &= \textsc{Transform}^l\Big( \mathbf{h}_i^{l-1},  \mathbf{h}_{\mathcal{N}_i}^l\Big),
    \label{eqn:GNN_update}
\end{aligned}
\end{equation}

\noindent where $\mathbf{h}_i^l$ is the latent representation of node $v_i$ at the $l$-th layer and $\mathcal{N}_i$ is the set of first-order neighboring nodes of node $v_i$. Notably, $\textsc{Aggregate}(\cdot)$ is an aggregation function that aggregates messages from neighboring nodes and $\textsc{Transform}(\cdot)$ computes the new representation of a node according to its previous-layer representation and the aggregated messages from neighbors.

% Specifically, the layer update process can be expressed in a matrix operation form. Given the input network, i.e., $\mathbf{G}=\{\mathbf{A},\mathbf{X}\}$, the representation updating of the $k$-th GNN layer follows the rule,

% \begin{equation}
%     \mathbf{H}^k = \sigma(\tilde{\mathbf{A}}\mathbf{H}^{k-1}\Theta_e^k)
% \end{equation}

% \noindent where $\mathbf{H}^k = [\mathbf{h}_1^k,\mathbf{h}_2^k,\cdots,\mathbf{h}_n^k]'\in\mathbb{R}^{n\times d_k}$ is the matrix form of the node representations at the $k$-th GNN layer, $\sigma$ represents pointwise nonlinear activation (e.g., ReLU), $\Theta_e^k$ is the weight matrix of the $k$-th GNN layer of the \textit{network encoder}~\kz{where $e$ denotes ...}, $\tilde{\mathbf{A}}=\tilde{\mathbf{D}}^{-\frac{1}{2}}\mathbf{A}_s\tilde{\mathbf{D}}^{-\frac{1}{2}}$ is the normalized adjacency matrix with self-loops where $\mathbf{A}_s =\mathbf{A}+\mathbf{I}$, $\tilde{\mathbf{D}}$ is the degree matrix of $\mathbf{A}_s$. Obviously,  $\mathbf{H}^0 = \mathbf{X}$ and we use $\mathbf{Z}=[\mathbf{z}_1,\mathbf{z}_2,\dots,\mathbf{z}_n]'\in\mathbb{R}^{n\times d_L}$ to denote the output representations from the \textit{network encoder} with $L$ GNN layers. For simplicity, we use the parameterized function $f_e(\cdot;\Theta_e):\mathbf{G}\mapsto\mathcal{Z}\in\mathbb{R}^{d_L}$ to represent the \textit{network encoder}. 

To capture the long-range node dependencies in the network, we stack multiple GNN layers in the \textit{network encoder}. Thus, the \textit{network encoder} can be represented by:
\begin{equation}
\begin{aligned}
    &\mathbf{H}^{1} = \text{GNN}^1 (\mathbf{A}, \mathbf{X}),\\
    &\dots\\
    &\mathbf{Z} = \text{GNN}^L (\mathbf{A}, \mathbf{H}^{L-1}),
\end{aligned}
\end{equation}
where $\mathbf{Z}$ is the learned node representations from the \textit{network encoder}. For simplicity, we use a parameterized function $f_{\bm\theta_e}(\cdot)$ to denote the \textit{network encoder} with $L$ GNN layers throughout the paper. It is worth noting that the \textit{network encoder} is compatible with arbitrary GNN-based architecture~\cite{kipf2017semi,hamilton2017inductive,velickovic2017graph,wu2019simplifying}, and here we employ Simple Graph Convolution (SGC)~\cite{wu2019simplifying} in our implementation.  

\smallskip
\noindent\textbf{Abnormality Valuator.} Afterwards, the learned node representations from the \textit{network encoder} will be passed to the \textit{abnormality valuator} $f_{\bm\theta_s}(\cdot)$ for further estimating the abnormality of each node. Specifically, the \textit{abnormality valuator} is built with two feed-forward layers that transform the intermediate node representations to scalar anomaly scores:
% \begin{equation}
%     f_s(\mathbf{z}_i;\Theta_s) =\mathbf{W}_2' \sigma(\mathbf{W}_1'\mathbf{z}_i)
%     \label{eqn:ano_scoring}
% \end{equation}

\begin{equation}
\begin{aligned}
    \mathbf{o}_i &= \text{ReLU}(\mathbf{W}_s \mathbf{z}_i + \mathbf{b}_s),\\
    s_i &= \mathbf{u}^{\mathrm{T}}_s \mathbf{o}_i + b_s,
\end{aligned}  
\label{eqn:ano_valuator}
\end{equation}
where $s_i$ is the anomaly score of node $v_i$ and $\mathbf{o}_i$ is the intermediate output. $\mathbf{W}_s$ and $\mathbf{u}_s$ are the learnable weight matrix and weight vector, respectively. $\mathbf{b}_s$ and $b_s$ are corresponding bias terms.

% Formally, we denote the \textit{anomaly scoring function} as $f_s(\cdot;\Theta_s):\mathcal{Z}\mapsto\mathcal{S}\in\mathbb{R}$, which converts the node representations with two feed-forward layers.

% \noindent where $\sigma$ denotes the nonlinear activation function, e.g., \relu, $\Theta_s=\{\mathbf{W}_1,\mathbf{W}_2\}$ represents the parameters of the \textit{anomaly scoring function}, $\mathbf{W}_1\in\mathbb{R}^{d_L\times d_h}$ is the weight matrix for the intermediate linear layer, and $\mathbf{W}_2\in\mathbb{R}^{d_h\times 1}$ is the weight vector of the output layer for computing the scores. 

To be more concrete, the whole GDN model $f_{\bm\theta}(\cdot)$ can be formally represented as:
\begin{equation}
    f_{\bm\theta}(\mathbf{A}, \mathbf{X}) = f_{\bm\theta_s}(f_{\bm\theta_e}(\mathbf{A}, \mathbf{X})),
\end{equation}
which directly maps the input network to scalar anomaly scores, and can be trained in an end-to-end fashion.

\smallskip
\noindent\textbf{Deviation Loss.} In essence, the objective of GDN is to distinguish normal and abnormal nodes according to the computed anomaly scores with few-shot labels. Here we propose to adopt the deviation loss~\cite{pang2019deep} to enforce the model to assign large anomaly scores to those nodes whose characteristics significantly deviate from normal nodes. To guide the model learning, we first define a \textit{reference score} (i.e., $\mu_r$) as the mean value of the anomaly scores of a set of randomly selected normal nodes. It serves as the reference to quantify how much the scores of anomalies deviate from those of normal nodes. 

According to previous studies~\cite{pang2019deep,kriegel2011interpreting}, Gaussian distribution is commonly a robust choice to fit the abnormality scores for a wide range of datasets. Based on this assumption, we first sample a set of $k$ anomaly scores from the Gaussian prior distribution, i.e., $\mathcal{R} = \{r_1, r_2, \dots, r_k\} \sim \mathcal{N}(\mu, \sigma^2)$, each of which denotes the abnormality of a random normal node. The \textit{reference score} is computed as the mean value of all the sampled scores:
\begin{equation}
    \mu_r = \frac{1}{k}\sum_{i=1}^k r_i .
    \label{eqn:reference_score}
\end{equation}

% intermediate node representation learning and node abnormality scoring is two-fold, (1) in the low-dimensional latent space, normal objects are confined in a specific cluster, which is determined by a prior distribution, while the anomalies are pushed far away from the cluster, and (2) regarding anomaly scores, nodes whose characteristics significantly deviate from those of normal nodes should be assigned large scores, which provides an effective approach to detect disparate anomalies from different auxiliary networks. Consequently, the deviation loss-based optimization enables to train an anomaly detector with remarkably less labeled data. 

% Following~\cite{pang2019deep}, we define the \textit{reference score}, i.e., $\mu_r$, which serves as the datum to quantify how much the scores of anomalies are different from those of normal objects, as the mean of the anomaly scores of a set of randomly selected normal objects. According to~\cite{kriegel2011interpreting}, for a range of datasets, the distribution of anomaly scores can be well approximated by Gaussian distribution, i.e., $s \sim \mathcal{N}(\mu, \sigma^2)$. To compute $\mu_r$, we first sample a set of $k$ anomaly scores from the Gaussian prior distribution, i.e., $s_1, s_2, \dots,s_k \sim \mathcal{N}(\mu_0, \sigma_0^2)$, according to the definition, the \textit{reference score} is computed as, 
% \begin{equation}
%     \mu_r = \frac{1}{k}\sum_{i=1}^ks_i
%     \label{eqn:reference_score}
% \end{equation}

With the \textit{reference score} $\mu_r$, the deviation between the anomaly score of node $v_i$ and the \textit{reference score} can be defined in the form of standard score:
\begin{equation}
\text{dev}(v_i) = \frac{s_i - \mu_r}{\sigma_r},
\label{eqn:dev}
\end{equation}
where $\sigma_r$ is the standard deviation of the set of sampled anomaly scores $\mathcal{R} = \{r_1,\dots,r_k\}$. Then the final objective function can be derived from the contrastive loss~\cite{hadsell2006dimensionality} by replacing the distance function with the deviation in Eq.~\eqref{eqn:dev}:
\begin{equation}
    \mathcal{L} = (1-y_i)\cdot|\text{dev}(v_i)| + y_i\cdot \text{max}(0, m-\text{dev}(v_i)),
    \label{eqn:dev_loss}
\end{equation}
where $y_i$ is the ground-truth label of input node $v_i$. If node $v_i$ is an abnormal node, $y_i=1$, otherwise, $y_i=0$. Note that $m$ is a confidence margin which defines a radius around the deviation.

By minimizing the above loss function, GDN will push the anomaly scores of normal nodes as close as possible to $\mu_r$ while enforcing a large positive deviation of at least $m$ between $\mu_r$ and the anomaly scores of abnormal nodes. This way GDN is able to learn a high-level abstraction of normal patterns with substantially less labeled anomalies, and empowers the node representation learning to discriminate normal nodes from the rare anomalies. Accordingly, a large anomaly score will be assigned to a node if its pattern significantly deviates from the learned abstraction of normal patterns.

% This offers an effective detection of dissimilar anomalies, e.g., anomalies due to different reasons or previously unknown anomalies; 
% and in turn the optimization also requires substantially less labeled anomalies to train the detector

% it not only optimizes the anomaly scores, but also empowers the intermediate representation learning to discriminate normal objects from the rare anomalies with different anomalous behaviors.

% The interpretation of the objective function is as follows, for normal nodes (i.e., $y=0$), the scoring function is encouraged to assign anomaly scores close to the \textit{reference score}, and for anomalous nodes (i.e., $y=1$), scores that are significantly larger than \textit{reference score} can dramatically minimize the loss. 
% , which encourages large positive deviations for all anomalies. the deviation loss is equivalent to enforcing statistically significant deviations of the scores of abnormal nodes from that of normal nodes in the upper tail. 

Our preliminary results show that GDN is not sensitive to the choices of $\mu$ and $\sigma$ as long as $\sigma$ is not too large. Specifically, we set $\mu = 0$ and $\sigma = 1$ in our experiments, which helps GDN to achieve stable detection performance on different datasets. It is also worth mentioning that, as we cannot access the labels of normal nodes, we simply consider the unlabeled node in $\mathcal{V}^U$ as normal. Note that this way the remaining unlabeled anomalies and all the normal nodes will be treated as normal, thus contamination is introduced to the training set (i.e., the ratio of unlabeled anomalies to the total unlabeled training data $\mathcal{V}^U$). Remarkably, GDN performs very well by using this simple strategy and is robust to different contamination levels. The effect of different contamination levels to model performance is evaluated in Sec.~\ref{subsec:ablation_robust}.

% As stated in Sec.~\ref{subsec:prob_defn}, for each auxiliary network $\mathbf{G}_i$, we have a set of few labeled anomalies.

%  Specifically, given the the total number of anomalies in the network, $a$ , the contamination ratio is computed as $r_c = \frac{a - |\mathcal{V}_i^L|}{|\mathcal{V}_i^U|}$ where $\mathcal{V}_i^L$ is the set of labeled anomalies.

% To summarize, GDN leverages a few labeled anomalies and the prior of anomaly scores to learn a high-level abstraction of normal patterns, enabling it to assign a large anomaly score to a node if its pattern significantly deviate from the learned abstraction.

\subsection{Cross-network Meta-learning}
Having the proposed Graph Deviation Networks (GDN), we are able to effectively detect anomalies on an arbitrary network with limited labeled data. When auxiliary networks from the same domain of the target network are available, how to transfer such valuable knowledge is the key to enable few-shot anomaly detection on the target network. Despite its feasibility, the performance would be rather limited if we directly borrow the idea of existing cross-network learning methods. The main reason is that those methods merely focus on transferring the knowledge from only a single network~\cite{wu2020unsupervised,shen2020adversarial}, which may cause negative transfer due to the divergent characteristics of anomalies on different networks. To this end, we turn to exploit multiple auxiliary networks to distill comprehensive knowledge of anomalies.

As an effective paradigm for extracting and transferring knowledge, meta-learning has recently received increasing research attention because of the broad applications in a variety of high-impact domains~\cite{santoro2016meta,vinyals2016matching,ding2020graph,wang2020graph,liu2019prototype,liu2021isometric}. In essence, the goal of meta-learning is to train a model on a variety of learning tasks, such that the learned model is capable of effectively adapting to new tasks with very few or even one labeled data~\cite{hochreiter2001learning}. In particular, Finn et al.~\cite{finn2017model} propose a model-agnostic meta-learning algorithm to explicitly learn the model parameters such that the model can achieve good generalization to a new task through a small number of gradient steps with limited labeled data. Inspired by this work, we propose to learn a meta-learner (i.e., Meta-GDN) as the initialization of GDN from multiple auxiliary networks, which possesses the generalization ability to effectively identify anomalous nodes on a new target network. Specifically, Meta-GDN extracts meta-knowledge of ground-truth anomalies from different few-shot network anomaly detection tasks on auxiliary networks during the training phase, and will be further fine-tuned for the new task on the target network, such that the model can make fast and effective adaptation.

% Formally, we use $\mathcal{L}_{\mathcal{T}_i}$ to denote the loss function of task $\mathcal{T}_i$ where $f_{\Theta}$ represents the graph deviation network that is parameterized by $\Theta$ (i.e., $f(\mathbf{G};\Theta)$) for simplicity. 

We define each learning task as performing few-shot anomaly detection on an individual network, whose objective is to enforce large anomaly scores to be assigned to anomalies as defined in Eq. (\ref{eqn:dev_loss}). Let $\mathcal{T}_i$ denote the few-shot network anomaly detection task constructed from network $\mathbf{G}_i^s$, then we have $P$ learning tasks in each epoch. We consider a GDN model represented by a parameterized function $f_{\bm\theta}$ with parameters $\bm\theta$. Given $P$ tasks, the optimization algorithm first adapts the initial model parameters $\bm\theta$ to $\bm\theta_i'$ for each learning task $\mathcal{T}_i$ independently. Specifically, the updated parameter $\bm\theta_i'$ is computed using $\mathcal{L}_{\mathcal{T}_i}$ on a batch of training data sampled from $\mathcal{V}^L_i$ and $\mathcal{V}^U_i$ in $\mathbf{G}^s_i$. Formally, the parameter update with one gradient step can be expressed as:
\begin{equation}
    \bm\theta_i' = \bm\theta - \alpha\nabla_{\bm\theta}\mathcal{L}_{\mathcal{T}_i}(f_{\bm\theta}),
    \label{eqn:update_theta}
\end{equation}
where $\alpha$ controls the meta-learning rate. Note that Eq.~\eqref{eqn:update_theta} only includes one-step gradient update, while it is straightforward to extend to multiple gradient updates~\cite{finn2017model}.

The model parameters are trained by optimizing for the best performance of $f_{\bm\theta}$ with respect to $\bm\theta$ across all learning tasks. More concretely, the meta-objective function is defined as follows:
\begin{equation}
\min_{\bm\theta} \sum_{i=1}^P \mathcal{L}_{\mathcal{T}_i}(f_{\bm\theta_i'}) = \min_{\bm\theta}\sum_{i=1}^P\mathcal{L}_{\mathcal{T}_i}(f_{\bm\theta - \alpha\nabla_{\bm\theta}\mathcal{L}_{\mathcal{T}_i}(f_{\bm\theta})}).
    \label{eqn:meta_loss}
\end{equation}

By optimizing the objective of GDN, the updated model parameter can preserve the capability of detecting anomalies on each network. Since the meta-optimization is performed over parameters $\bm\theta$ with the objective computed using the updated parameters (i.e., $\bm\theta_i'$) for all tasks, correspondingly, the model parameters are optimized such that one or a small number of gradient steps on the target task (network) will produce great effectiveness.

Formally, we leverage stochastic gradient descent (SGD) to update the model parameters $\bm\theta$ across all tasks, such that the model parameters $\bm\theta$ are updated as follows:
\begin{equation}
    \bm\theta \leftarrow \bm\theta - \beta\nabla_{\bm\theta}\sum_{i=1}^P\mathcal{L}_{\mathcal{T}_i}(f_{\bm\theta_i'}),
\end{equation}
where $\beta$ is the meta step size. The full algorithm is summarized in Algorithm~\ref{alg:training}. Specifically, for each batch, we randomly sample the same number of nodes from unlabeled data (i.e., $\mathcal{V}^U$) and labeled anomalies (i.e., $\mathcal{V}^L$) to represent normal and abnormal nodes, respectively (Step-\ref{alg:sample_batch}).

\begin{algorithm}[!t]
\caption{The learning algorithm of Meta-GDN}\label{alg:training}
\begin{algorithmic}[1]
\Require (1) $P$ auxiliary networks, i.e., $\mathcal{G}^s = \{\mathbf{G}_1^s = (\mathbf{A}_1^s, \mathbf{X}_1^s),\mathbf{G}_2^s = (\mathbf{A}_2^s, \mathbf{X}_2^s),  \dots,\mathbf{G}_P^s =(\mathbf{A}_P^s, \mathbf{X}_P^s) \}$; (2) a target network $\mathbf{G}^t = (\mathbf{A}^t, \mathbf{X}^t)$; (3) sets of few-shot labeled anomalies and unlabeled nodes for each network (i.e., $\{\mathcal{V}_1^L, \mathcal{V}_1^U\}, \dots, \{\mathcal{V}_P^L, \mathcal{V}_P^U\}$ and $\{\mathcal{V}_t^L, \mathcal{V}_t^U\}$); (4) training epochs $E$, batch size $b$, and meta-learning hyper-parameters $\alpha, \beta$.
% (2) target network $\mathbf{G}^t$ with labeled anomalies $\mathcal{V}_t^L$; 

\Ensure Anomaly scores of nodes in $\mathcal{V}_t^U$.
\State Initialize parameters $\bm{\theta}$;
% \State Sample a batch of networks from $\mathcal{G}^s$;
\While{$e<E$}
\For{each network $\mathbf{G}_i^s$ (task $\mathcal{T}_i$)}
\State \multiline{Randomly sample $\frac{b}{2}$ nodes from $\mathcal{V}_i^L$ and $\frac{b}{2}$ from $\mathcal{V}_i^U$ to comprise the batch $B_i$;}\label{alg:sample_batch}
% with half objects from $\mathcal{V}_i^L$ and another half from the rest of $\mathbf{G}_i$;
\State Evaluate  $\nabla_{\bm{\theta}}\mathcal{L}_{\mathcal{T}_i}(f_{\bm{\theta}})$ using $B_i$ and $\mathcal{L}(\cdot)$ in Eq.~\eqref{eqn:dev_loss};
\State \multiline{%
Compute adapted parameters $\bm{\theta}'$ with gradient descent using Eq.~\eqref{eqn:update_theta}, $\bm{\theta}_i' \leftarrow \bm{\theta} - \alpha\nabla_{\bm{\theta}}\mathcal{L}_{\mathcal{T}_i}(f_{\bm{\theta}})$;} 
\State \multiline{%
Sample a new batch $B_i'$ for the meta-update;} 
\EndFor
\State \multiline{Update $\bm{\theta} \leftarrow \bm{\theta} - \beta\nabla_{\bm{\theta}}\sum_{i=1}^p\mathcal{L}_{\mathcal{T}_i}(f_{\bm{\theta}_i'})$ using $\{B_i'\}$ and $\mathcal{L}(\cdot)$ according to Eq.~\eqref{eqn:dev_loss};}
\EndWhile
\State Fine-tune $\bm{\theta}$ on target network $\mathbf{G}^t$ with $\{\mathcal{V}_t^L, \mathcal{V}_t^U\}$;
\State Compute anomaly scores for nodes in $\mathcal{V}_t^U$;
\end{algorithmic}
\end{algorithm}

% New devnet structure: (1) scoring network, $\{f_e, f_s\}$, where $f_e:\mathbb{R}^{n\times d}\rightarrow\mathbb{R}^{n\times l}$ is a gnn based network that learns latent representations for nodes; $f_s:\mathbb{R}^{n\times l}\rightarrow\mathbb{R}^{n\times 1}$ is an anomaly scoring function which scores each node in the graph.
\section{Experiments}
\label{sec:exp}
In this section, we perform empirical evaluations to demonstrate the effectiveness of the proposed framework. Specifically, we aim to answer the following research questions:

\begin{itemize}
    \item \textbf{RQ1.} How effective is the proposed approach Meta-GDN for detecting anomalies on the target network with few or even one labeled instance? 
    \item \textbf{RQ2.} How much will the performance of Meta-GDN change by providing different numbers of auxiliary networks or different anomaly contamination levels?
    \item \textbf{RQ3.} How does each component of Meta-GDN (i.e., graph deviation networks or cross-network meta-learning) contribute to the final detection performance?
\end{itemize}
\subsection{Experimental Setup}\label{subsec:exp_setup}

\noindent\textbf{Evaluation Datasets.}
In the experiment, we adopt three real-world datasets, which are publicly available and have been widely used in previous research~\cite{rayana2015collective,sen2008collective,kipf2017semi,hamilton2017inductive}. Table~\ref{tab:data_stats} summarizes the statistics of each dataset. The detailed description is as follows:

%\cite{rayana2015collective5,kipf2017semi,DBLP:conf/nips/HamiltonYL17}.
\begin{itemize}
    \item \textbf{Yelp}~\cite{rayana2015collective} is collected from Yelp.com and contains reviews for restaurants in several states of the U.S., where the restaurants are organized by ZIP codes. The reviewers are classified into two classes, abnormal (reviewers with only filtered reviews) and normal (reviewers with no filtered reviews) according to the Yelp anti-fraud filtering algorithm. We select restaurants in the same location according to ZIP codes to construct each network, where nodes represent reviewers and there is a link between two reviewers if they have reviewed the same restaurant. We apply the bag-of-words model~\cite{zhang2010understanding} on top of the textual contents to obtain the attributes of each node. 
    
    % The vocabulary is built on top of the textual contents.
    \item \textbf{PubMed}~\cite{sen2008collective} is a citation network where nodes represent scientific articles related to diabetes and edges are citations relations. Node attribute is represented by a TF/IDF weighted word vector from a dictionary which consists of 500 unique words. We randomly partition the large network into non-overlapping sub-networks of similar size.
    \item \textbf{Reddit}~\cite{hamilton2017inductive} is collected from an online discussion forum where nodes represent threads and an edge exits between two threads if they are commented by the same user. The node attributes are constructed using averaged word embedding vectors of the threads. Similarly, we extract non-overlapping sub-networks from the original large network for our experiments.
\end{itemize}

\begin{table}[!t]
	\centering
	  \caption{Statistics of evaluation datasets. $r_1$ denotes the ratio of labeled anomalies to the total anomalies and $r_2$ is the ratio of labeled anomalies to the total number of nodes.}
	{\begin{tabular}{lccc}
% 	\headrulewidth
    \toprule
% 	\hline
    \textbf{Datasets} & Yelp & PubMed & Reddit \\
    \midrule
    \# nodes (avg.) & $4,872$ & $3,675$ & $15,860$ \\
    \# edges (avg.) & $43,728$ & $8,895$ & $136,781$ \\
    \# features & $10,000$ & $500$ & $602$ \\
    \# anomalies (avg.) & $223$ & $201$ & $796$\\
    $r_1$ (avg.) & $4.48\%$ & $4.97\%$ & $1.26\%$ \\
    $r_2$ (avg.) & $0.21\%$ & $0.27\%$ & $0.063\%$ \\
%     \textbf{Datasets} & \# nodes (avg.) & \# edges (avg.) &  \# features & \# anomalies (avg.) \\
% 	\midrule
% 	Yelp & $4,872$ & $43,728$ & $10,000$ & $223$ \\
%     Pubmed & $3,675$ & $8,895$ & $500$ & $201$ \\
%     Reddit & $15,860$ & $136,781$ & $602$ & $796$ \\
    \bottomrule
	\end{tabular}
	}
% 	\vspace{-5mm}
  
	\label{tab:data_stats}
\end{table}

Note that except the \textit{Yelp} dataset, we are not able to access ground-truth anomalies for \textit{PubMed} and \textit{Reddit}. Thus we refer to two anomaly injection methods~\cite{song2007conditional,ding2019interactive} to inject a combined set of anomalies (i.e., structural anomalies and contextual anomalies) by perturbing the topological structure and node attributes of the original network, respectively. To inject structural anomalies, we adopt the approach used by~\cite{ding2019interactive} to generate a set of small cliques since small clique is a typical abnormal substructure in which a small set of nodes are much more closely linked to each other than average~\cite{skillicorn2007detecting}. Accordingly, we randomly select $c$ nodes (i.e., clique size) in the network and then make these nodes fully linked to each other. By repeating this process $K$ times (i.e., $K$ cliques), we can obtain $K \times c$ structural anomalies. In our experiment, we set the clique size $c$ to $15$. In addition, we leverage the method introduced by ~\cite{song2007conditional} to generate contextual anomalies. Specifically, we first randomly select a node $i$ and then randomly sample another 50 nodes from the network. We choose the node $j$ whose attributes have the largest Euclidean distance from node $i$ among the 50 nodes. The attributes of node $i$ (i.e., $\mathbf{x}_i$) will then be replaced with the attributes of node $j$ (i.e., $\mathbf{x}_j$). Note that we inject structural and contextual anomalies with the same quantity and the total number of injected anomalies is around $5\%$ of the network size.

\smallskip
\noindent\textbf{Comparison Methods.} We compare our proposed Meta-GDN framework and its base model GDN with two categories of anomaly detection methods, including (1) \textit{feature-based} methods (i.e., LOF, Autoencoder and DeepSAD) where only the node attributes are considered, and (2) \textit{network-based} methods (i.e., SCAN, ConOut, Radar, DOMINANT, and SemiGNN) where both topological information and node attributes are involved. Details of these compared baseline methods are as follows:

\begin{itemize}
    \item \textbf{LOF}~\cite{breunig2000lof} is a feature-based approach which detects outliers at the contextual level.
    \item \textbf{Autoencoder}~\cite{zhou2017anomaly} is a feature-based unsupervised deep autoencoder model which introduces an anomaly regularizing penalty based upon L1 or L2 norms.
    % , which learns data embeddings by minimizing the difference between the original input features and the output\hide{\kz{Anomaly Detection with Robust Deep Autoencoders}}. To evaluate, the anomaly score is defined as the aggregation of the values from the corresponding embedding\kz{rewrite this more to be more concise}.
    \item \textbf{DeepSAD}~\cite{ruff2019deep} is a state-of-the-art deep learning approach for general semi-supervised anomaly detection. In our experiment, we leverage the node attribute as the input feature.
    % \item \textbf{DevNet} is an end-to-end feature-based approach which aims to leverage very limited labeled data to detect anomalies. The key idea is to assign significantly larger scores to anomalous objects than normal ones~\cite{pang2019deep}.
    \item \textbf{SCAN}~\cite{xu2007scan} is an efficient algorithm for detecting network anomalies based on a structural similarity
    measure.
    \item \textbf{ConOut}~\cite{sanchez2014local} identifies network anomalies according to the corresponding subgraph and the relevant subset of attributes in the local context.
    \item \textbf{Radar}~\cite{li2017radar} is an unsupervised method that detects anomalies on attributed network by characterizing the residuals of attribute information and its coherence with network structure.
    \item \textbf{DOMINANT}~\cite{ding2019deep} is a GCN-based autoencoder framework which computes anomaly scores using the reconstruction errors from both network structure and node attributes.
    \item \textbf{SemiGNN}~\cite{wang2019semi} is a semi-supervised GNN model, which leverages the hierarchical attention mechanism to better correlate different neighbors and different views.

\end{itemize}

% \noindent\textbf{C - Parameter Setting.}

\begin{table*}[!ht]
\centering
\caption{Performance comparison results (10-shot) w.r.t. AUC-ROC and AUC-PR on three datasets.}
\scalebox{1.0}{
\begin{tabular}{@{}ll|cccccc@{}}
\toprule
\multicolumn{2}{l||}{} & \multicolumn{2}{c}{Yelp} & \multicolumn{2}{c}{PubMed} & \multicolumn{2}{c}{Reddit} \\ \cmidrule{3-8}
\multicolumn{2}{l||}{\textbf{Methods}} & \multicolumn{1}{c|}{AUC-ROC} & \multicolumn{1}{c|}{AUC-PR} & \multicolumn{1}{c|}{AUC-ROC} & \multicolumn{1}{c|}{AUC-PR} & \multicolumn{1}{c|}{AUC-ROC} & \multicolumn{1}{c}{AUC-PR}   \\ 
\midrule \midrule
\multicolumn{2}{l||}{LOF} 
& $0.375\pm0.011$ &  \multicolumn{1}{c|}{$0.042\pm0.004$} & $0.575\pm0.007$
& \multicolumn{1}{c|}{$0.187\pm0.016$} & $0.518\pm0.015$ & $0.071\pm0.006$
\\
\multicolumn{2}{l||}{Autoencoder} 
& $0.365\pm0.013$ & \multicolumn{1}{c|}{$0.041\pm0.008$} & $0.584\pm0.018$
& \multicolumn{1}{c|}{$0.236\pm0.005$} & $0.722\pm0.012$ & $0.347\pm0.007$ 
\\
\multicolumn{2}{l||}{DeepSAD} 
& $0.460\pm0.008$ &  \multicolumn{1}{c|}{$0.062\pm0.005$} & $0.528\pm0.008$
& \multicolumn{1}{c|}{$0.115\pm0.004$}  &  $0.503\pm0.010$ &  $0.066\pm0.005$
% \multicolumn{2}{l||}{DevNet} 
% & $0.674\pm0.021$ &  \multicolumn{1}{c|}{$0.125\pm0.017$} & $0.687\pm0.013$
% & \multicolumn{1}{c|}{$0.384\pm0.019$}  &  $0.722\pm0.014$ &  $0.367\pm0.013$
\\
\midrule
\multicolumn{2}{l||}{SCAN}
& $0.397\pm0.011$ & \multicolumn{1}{c|}{$0.046\pm0.005$} & $0.421\pm0.016$
& \multicolumn{1}{c|}{$0.048\pm0.005$} & $0.298\pm0.009$ & $0.048\pm0.002$ 
\\
\multicolumn{2}{l||}{ConOut}
& $0.402\pm0.015$ & \multicolumn{1}{c|}{$0.041\pm0.005$} & $0.511\pm0.019$
& \multicolumn{1}{c|}{$0.093\pm0.012$} & $0.551\pm0.008$ & $0.085\pm0.007$  
\\
\multicolumn{2}{l||}{Radar}  
& $0.415\pm0.012$ &  \multicolumn{1}{c|}{$0.045\pm0.007$} & $0.573\pm0.013$
& \multicolumn{1}{c|}{$0.244\pm0.011$} & $0.721\pm0.008$ & $0.281\pm0.007$ 
\\
\multicolumn{2}{l||}{DOMINANT}
& $0.578\pm0.018$ & \multicolumn{1}{c|}{$0.109\pm0.003$} & $0.636\pm0.021$
& \multicolumn{1}{c|}{$0.337\pm0.013$} & $0.735\pm0.013$ & $0.357\pm0.009$  
\\
\multicolumn{2}{l||}{SemiGNN}
& $0.497\pm0.004$ & \multicolumn{1}{c|}{$0.058\pm0.003$} & $0.523\pm0.008$
& \multicolumn{1}{c|}{$0.065\pm0.006$} & $0.610\pm0.007$ & $0.134\pm0.003$ 
\\
\midrule
\multicolumn{2}{l||}{GDN (ours)} 
& $0.678\pm0.015$ & \multicolumn{1}{c|}{$0.132\pm0.009$} & $0.736\pm0.012$
& \multicolumn{1}{c|}{$0.438\pm0.012$} & $0.811\pm0.015$ & $0.379\pm0.011$
\\
% \multicolumn{2}{l||}{GDN\textsuperscript{+} (ours)} 
% & $0.691\pm 0.020$ & \multicolumn{1}{c|}{$0.141\pm 0.011$} & $0.756\pm 0.031$
% & \multicolumn{1}{c|}{$0.516\pm0.028$} & $0.860\pm0.029$ & $0.408\pm0.014$
% \\
\multicolumn{2}{l||}{Meta-GDN (ours)} 
& $\mathbf{0.724\pm0.012}$ &  \multicolumn{1}{c|}{$\mathbf{0.175\pm0.011}$} & $\mathbf{0.761\pm0.014}$ & \multicolumn{1}{c|}{$\mathbf{0.485\pm0.010}$} & $\mathbf{0.842\pm0.011}$ & $\mathbf{0.395\pm0.009}$
\\ 
\bottomrule
\end{tabular}}

\label{tab:result_auc}
\end{table*}

\begin{figure*}[!ht]
    \centering
\scalebox{0.9}{
    \includegraphics[width=\textwidth]{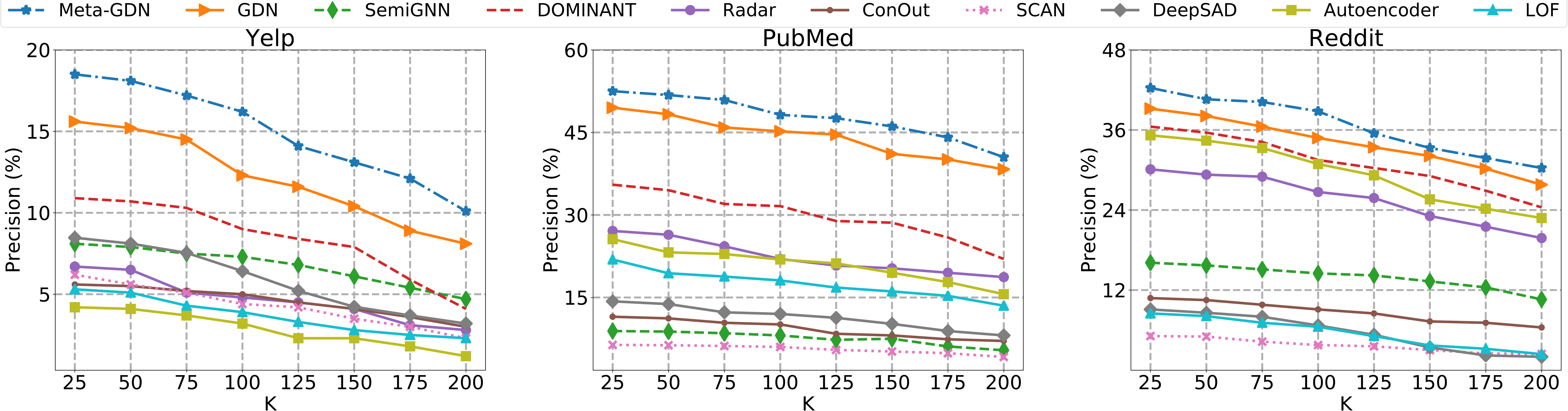}}
    \caption{Performance comparison results (10-shot) w.r.t. Precision@K on three datasets. Figure best viewed in color.}
    \label{fig:result_precisionatK}
\end{figure*}

\noindent\textbf{Evaluation Metrics.}
In this paper, we use the following metrics to have a comprehensive evaluation of the performance of different anomaly detection methods:

\begin{itemize}
    \item \textbf{AUC-ROC} is widely used in previous anomaly detection research~\cite{ding2019deep,li2017radar}. Area under curve (AUC) is interpreted as the probability that a randomly chosen anomaly receives a higher score than a randomly chosen normal object.
    % Receiver operating characteristic (ROC) curve is created by plotting the true positive rate (TPR, an anomaly is correctly detected) against the false positive rate (FPR, a normal object is incorrectly recognized as an anomaly) at a variety of threshold settings. 
    \item \textbf{AUC-PR} is the area under the curve of precision against recall at different thresholds, and it only evaluates the performance on the positive class (i.e., abnormal objects). AUC-PR is computed as the average precision as defined in~\cite{manning2008introduction} and is used as the evaluation metric in ~\cite{pang2019deep}.
    \item \textbf{Precision$\mathbf{@K}$} is defined as the proportion of true anomalies in a ranked list of $K$ objects. We obtain the ranking list in descending order according to the anomaly scores that are computed from a specific anomaly detection algorithm.
\end{itemize}

\noindent\textbf{Implementation Details.} Regarding the proposed GDN model, we use Simple Graph Convolution~\cite{wu2019simplifying} to build the \textit{network encoder} with degree $K=2$ (two layers). As shown in Eq.~\eqref{eqn:ano_valuator}, the \textit{abnormality valuator} employs a two-layer neural network with one hidden layer of $512$ units followed by an output layer of $1$ unit. The confidence margin (i.e., $m$) in Eq.~\eqref{eqn:dev_loss} is set as $5$ and the reference score (i.e., $\mu_r$) is computed using Eq.~\eqref{eqn:reference_score} from $k=5,000$ scores that are sampled from a Gaussian prior distribution, i.e., $\mathcal{N}(0,1)$. Unless otherwise specified, we set the total number of networks as $5$ ($4$ auxiliary networks and $1$ target network), and for each one we have access to $10$ labeled abnormal nodes that are randomly selected from the set of labeled anomalies ($\mathcal{V}^L$) in every run of the experiment. 

% Specifically, the unlabeled data ($\mathcal{U}$), which includes the remaining unlabeled anomalies and all the normal nodes, will be treated as normal.
For model training, the proposed GDN and Meta-GDN are trained with $1000$ epochs, with batch size $16$ in each epoch, and a $5$-step gradient update is leveraged to compute $\bm\theta'$ in the meta-optimization process. The network-level learning rate $\alpha$ is $0.01$ and the meta-level learning rate $\beta = 0.001$. Fine-tuning is performed on the target network where the corresponding nodes are split into $40\%$ for fine-tuning, $20\%$ for validation, and $40\%$ for testing. For all the comparison methods, we select the hyper-parameters with the best performance on the validation set and report the results on the test data of the target network for a fair comparison. Particularly, for all the network-based methods, the whole network structure and node attributes are accessible during training.

% We fine-tune the parameters of other comparison methods from the open-source implementations on these datasets.

% We also evaluate the performance of Meta-GDN given the following numbers of labeled anomalies on the target network, i.e., $1, 3, 5, 10$ while the number of labeled data remains the same on auxiliary networks. The results are presented in Sec.~\ref{subsec:effectiveness}.

% Parameter analysis on $p$ and $r_l$ is summarized in Sec.~\ref{subsec:parameter}. To investigate the robustness of GDN and Meta-GDN, we evaluate the influence of the contamination level (i.e., $r_c$) to the model performance on \textit{Reddit} dataset, where we fix the number of labeled anomalies to $10$ and increase the number of unlabeled anomalies by injecting two types of anomalies as described in Sec.~\ref{subsec:exp_setup}. We report the average results of 20 runs for all experiments.

\begin{table*}[!ht]
\centering
\caption{Few-shot performance evaluation of Meta-GDN w.r.t. AUC-ROC and AUC-PR.}
\scalebox{1.0}{
\begin{tabular}{@{}ll|cccccc@{}}
\toprule
\multicolumn{2}{l||}{} & \multicolumn{2}{c}{Yelp} & \multicolumn{2}{c}{PubMed} & \multicolumn{2}{c}{Reddit} \\ \cmidrule{3-8}
\multicolumn{2}{l||}{\textbf{Setting}} & \multicolumn{1}{c|}{AUC-ROC} & \multicolumn{1}{c|}{AUC-PR} & \multicolumn{1}{c|}{AUC-ROC} & \multicolumn{1}{c|}{AUC-PR} & \multicolumn{1}{c|}{AUC-ROC} & \multicolumn{1}{c}{AUC-PR}   \\ 
\midrule \midrule
\multicolumn{2}{l||}{$1$-shot} 
& $0.702\pm0.008$ &  \multicolumn{1}{c|}{$0.159\pm0.015$} & $0.742\pm0.012$ & \multicolumn{1}{c|}{$0.462\pm0.013$} & $0.821\pm0.013$ & $0.380\pm0.011$ 
\\
\multicolumn{2}{l||}{$3$-shot} 
& $0.709\pm0.006$ &  \multicolumn{1}{c|}{$0.164\pm0.010$} & $0.748\pm0.008$ & \multicolumn{1}{c|}{$0.468\pm0.008$} & $0.828\pm0.012$ & $0.386\pm0.007$ 
\\
\multicolumn{2}{l||}{$5$-shot} 
& $0.717\pm0.013$ &  \multicolumn{1}{c|}{$0.169\pm0.007$} & $0.753\pm0.011$ & \multicolumn{1}{c|}{$0.474\pm0.005$} & $0.834\pm0.009$ & $0.389\pm0.008$ 
\\
\multicolumn{2}{l||}{$10$-shot} 
& $0.724\pm0.012$ &  \multicolumn{1}{c|}{$0.175\pm0.011$} & $0.761\pm0.014$ & \multicolumn{1}{c|}{$0.485\pm0.010$} & $0.842\pm0.011$ & $0.395\pm0.009$
\\
\bottomrule
\end{tabular}}

\label{tab:result_auc_shot}
\end{table*}

\subsection{Effectiveness Results (RQ1)}\label{subsec:effectiveness}

% In addition, we also compare GDN in a scenario where the number of labeled anomalies in an individual network is approximately the same as the total number of labeled anomalies from all auxiliary networks for Meta-GDN, denoted as $\mathrm{GDN}^+$. 

\noindent\textbf{Overall Comparison.} In the experiments, we evaluate the performance of the proposed framework Meta-GDN along with its base model GDN by comparing with the included baseline methods. We first present the evaluation results (10-shot) w.r.t. AUC-ROC and AUC-PR in Table~\ref{tab:result_auc} and the results w.r.t. Precision@K are visualized in Figure~\ref{fig:result_precisionatK}. Accordingly, we have the following observations, including: \textbf{(1)} in terms of AUC-ROC and AUC-PR, our approach Meta-GDN outperforms all the other 
compared methods by a significant margin. Meanwhile, the results w.r.t. Precision@K again demonstrate that Meta-GDN can better rank abnormal nodes on higher positions than other methods by estimating accurate anomaly scores; \textbf{(2)} unsupervised methods (e.g., DOMINANT, Radar) are not able to leverage supervised knowledge of labeled anomalies and therefore have limited performance. Semi-supervised methods (e.g., DeepSAD, SemiGNN) also fail to deliver satisfactory results. The possible explanation is that DeepSAD cannot model network information and SemiGNN requires a relatively large number of labeled data and multi-view data, which make them less effective in our evaluation; and \textbf{(3)} compared to the base model GDN, Meta-GDN is capable of extracting comprehensive meta-knowledge across multiple auxiliary networks by virtue of the cross-network meta-learning algorithm, which further enhances the detection performance on the target network.

% \hide{For example, Meta-GDN improves GDN model by $3.6\%, 4.2\%$ and $1.6\%$ from the perspective of AUC-PR on three datasets, respectively,}

% \hide{(\qh{can be deleted} For example, on \textit{PubMed} dataset, GDN achieves $0.736$ and $0.438$ in AUC-ROC and AUC-PR, respectively, which is $10\%$ better than the best competitor, i.e., DOMINANT)}

\smallskip
\noindent\textbf{Few-shot Evaluation.} In order to verify the effectiveness of Meta-GDN in few-shot as well as one-shot network anomaly detection, we evaluate the performance of Meta-GDN with different numbers of labeled anomalies on the target network (i.e., $1$-shot, $3$-shot, $5$-shot and $10$-shot). Note that we respectively set the batch size $b$ to $2$, $4$, $8$, and $16$ to ensure that there is no duplication of labeled anomalies exist in a sampled training batch. Also, we keep the number of labeled anomalies on auxiliary networks as $10$. Table~\ref{tab:result_auc_shot} summarizes the AUC-ROC/AUC-PR performance of Meta-GDN under different few-shot settings. 
By comparing the results in Table~\ref{tab:result_auc} and Table~\ref{tab:result_auc_shot}, we can see that even with only one labeled anomaly on the target network (i.e., $1$-shot), Meta-GDN can still achieve good performance and significantly outperforms all the baseline methods. In the meantime, we can clearly observe that the performance of Meta-GDN increases with the growth of the number of labeled anomalies, which demonstrates that Meta-GDN can be better fine-tuned on the target network with more labeled examples.

% For example, while the $1$-shot Meta-GDN falling behind itself under the $10$-shot setting, it can significantly outperform the the best performing baseline, i.e., DOMINANT, under the $10$-shot setting in each dataset.

%Additionally, compared to the best performing baseline, i.e., DOMINANT, on \textit{PubMed} dataset, 1-shot still achieves $0.106$ better in terms of AUC-ROC. By providing more labeled anomalies, Meta-GDN is able to be 

% Similar results can be observed in Precision@K results, we omit that due to the space limit.

%AUC-PR of $1$-shot by Meta-GDN can outperform the best performing baseline, i.e., DOMINANT, 
% For example, the AUC-PR of $1$-shot is $9.14\%$, $4.74\%$ and $3.80\%$ less than that of $10$-shot setting, respectively.

% \kz{Compared to the best performing baseline, 1-shot....}
% 

% and (3) on \textit{PubMed} and \textit{Reddit} datasets, $\mathrm{GDN}^+$ achieves better performance than Meta-GDN w.r.t. AUC-ROC and AUC-PR, since $\mathrm{GDN}^+$ has more labeled anomalies in the target network than Meta-GDN, and characteristics of these labeled anomalies are similar because they are manually generated by certain approach. However, on \textit{Yelp} dataset, Meta-GDN outperforms $\mathrm{GDN}^+$, and an possible explanation is that the anomalies in this dataset may have very disparate behaviors and the information from the auxiliary networks provides extended knowledge about anomalies to train a more comprehensive and effective model.

% 1. auc\_roc, auc\_pr
% maml $\sim=$ g-dev with the same \#anomalies
% precision\@ K: line curve

\begin{figure}[h]
    \graphicspath{{figures/}}
    \centering
   
    \subfigure[]
    {
    \includegraphics[width=.45\columnwidth]{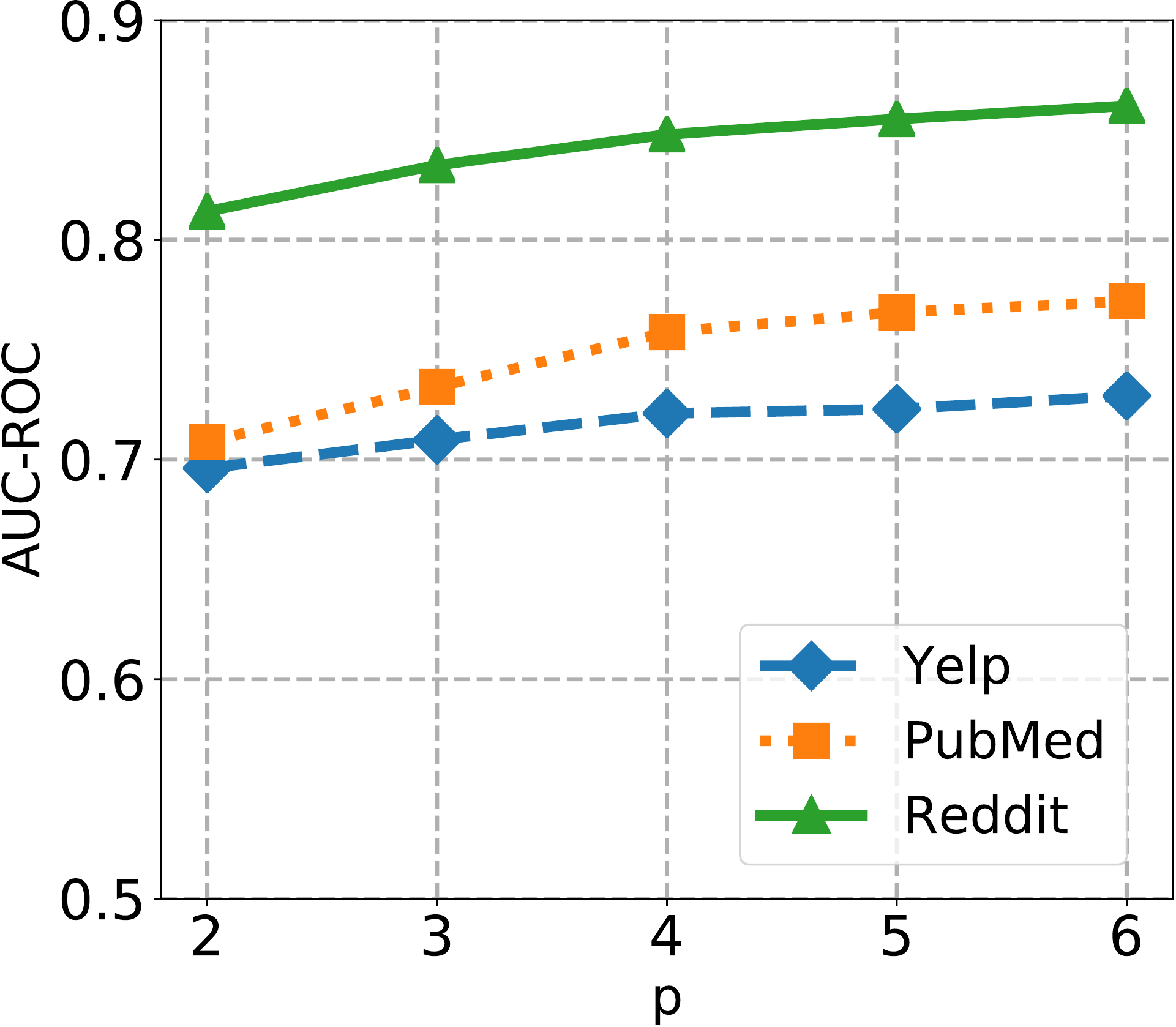}\label{fig:param_roc_p}
    }
    \hspace{0.1cm}
    \subfigure[]
    {
    \includegraphics[width=.45\columnwidth]{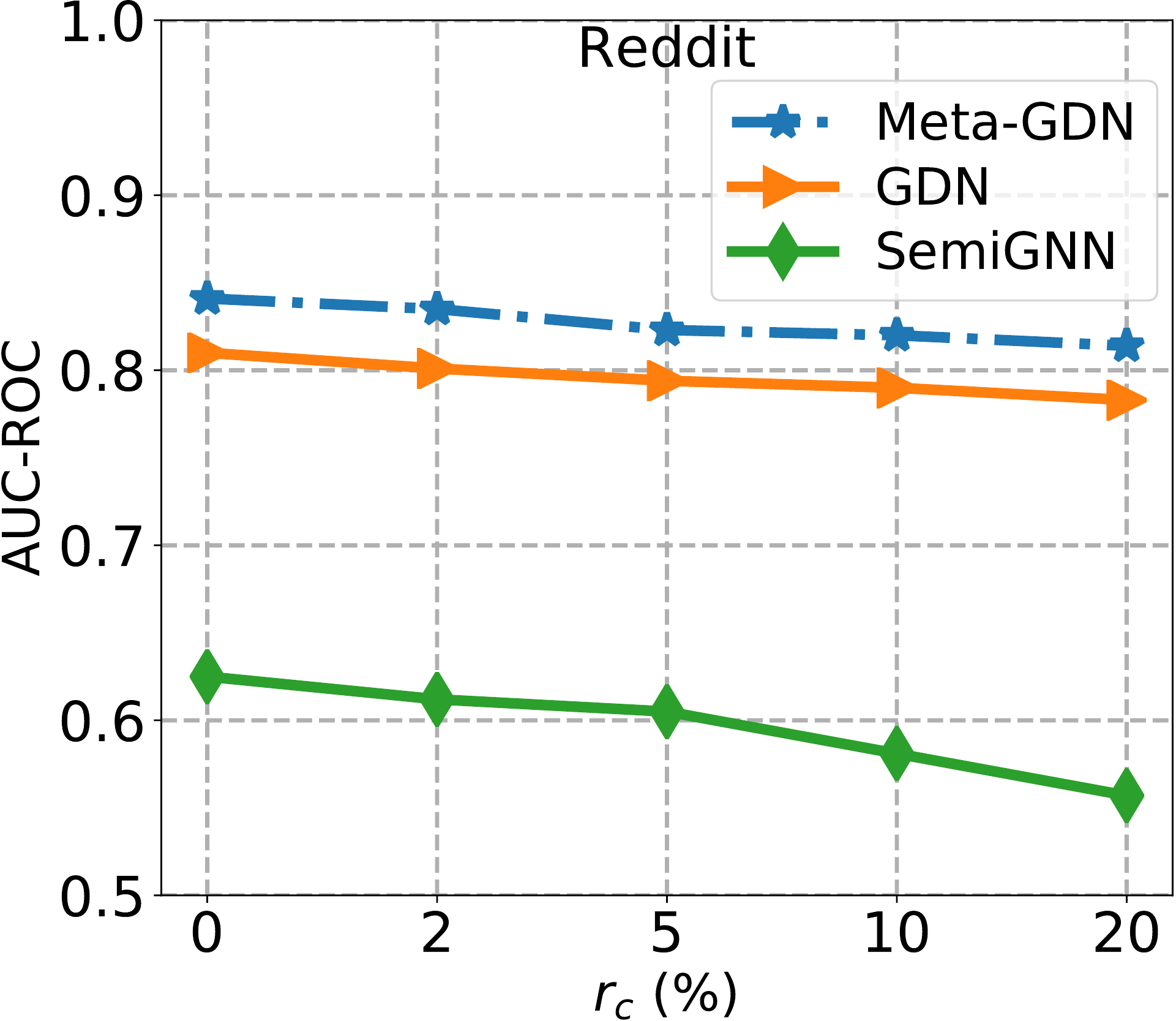}\label{fig:robust}
    }
     \caption{(a) Sensitivity analysis of Meta-GDN w.r.t. different number of auxiliary networks; (b) Model robustness study w.r.t. AUC-ROC with different contamination levels.}
    \label{fig:param_analysis}
\end{figure}

\subsection{Sensitivity \& Robustness Analysis (RQ2)}\label{subsec:parameter}

In this section, we further analyze the sensitivity and robustness of the proposed framework Meta-GDN. By providing different numbers of auxiliary networks during training, the model sensitivity results w.r.t. AUC-ROC are presented in Figure~\ref{fig:param_roc_p}. Specifically, we can clearly find that \textbf{(1)} as the number of auxiliary networks increases, Meta-GDN achieves constantly stronger performance on all the three datasets. It shows that more auxiliary networks can provide better meta-knowledge during the training process, which is consistent with our intuition; \textbf{(2)} Meta-GDN can still achieve relatively good performance when training with a small number of auxiliary networks (e.g., $p=2$), which demonstrates the strong capability of its base model GDN. For example, on \textit{Yelp} dataset, the performance barely drops $0.033$ if we change the number of auxiliary networks from $p=6$ to $p=2$.

As discussed in Sec.~\ref{subsec:gdn}, we treat all the sampled nodes from unlabeled data as normal for computing the deviation loss. This simple strategy introduces anomaly contamination in the unlabeled training data. Due to the fact that $r_c$ is a small number in practice, our approach can work very well in a wide range of real-world datasets. To further investigate the robustness of Meta-GDN w.r.t. different contamination levels $r_c$ (i.e., the proportion of anomalies in the unlabeled training data), we report the evaluation results of Meta-GDN, GDN and the semi-supervised baseline method SemiGNN in Figure~\ref{fig:robust}. As shown in the figure, though the performance of all the methods decreases with increasing
contamination levels, both Meta-GDN and GDN are remarkably robust and can consistently outperform SemiGNN to a large extent. 

\begin{figure}[!t]
    \graphicspath{{figures/}}
    \centering
    \subfigure[]
    {
     \includegraphics[width=0.45\columnwidth]{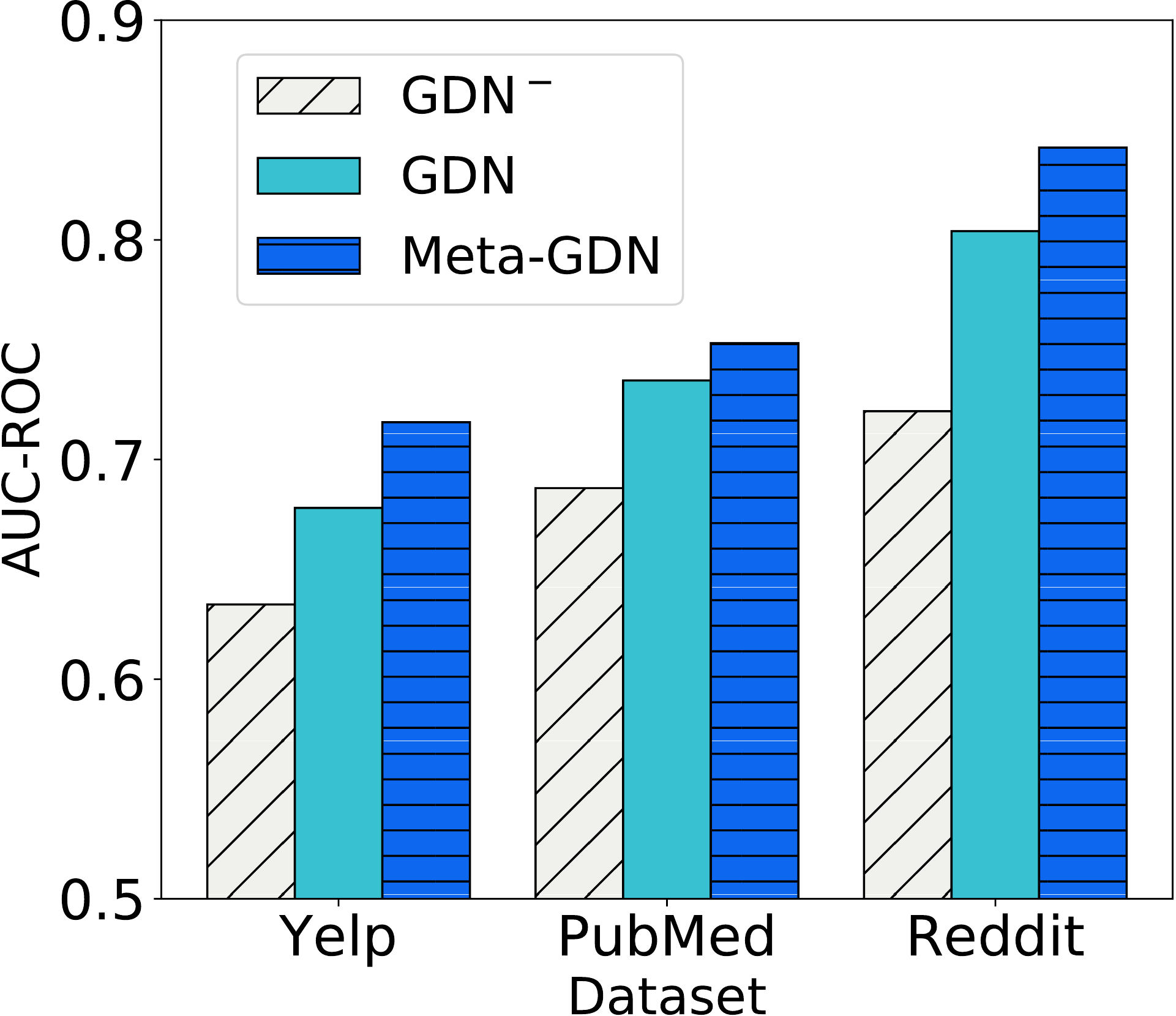}\label{fig:ablation_roc}
    }
    \hspace{0.1cm}
    \subfigure[]
    {

     \includegraphics[width=0.45\columnwidth]{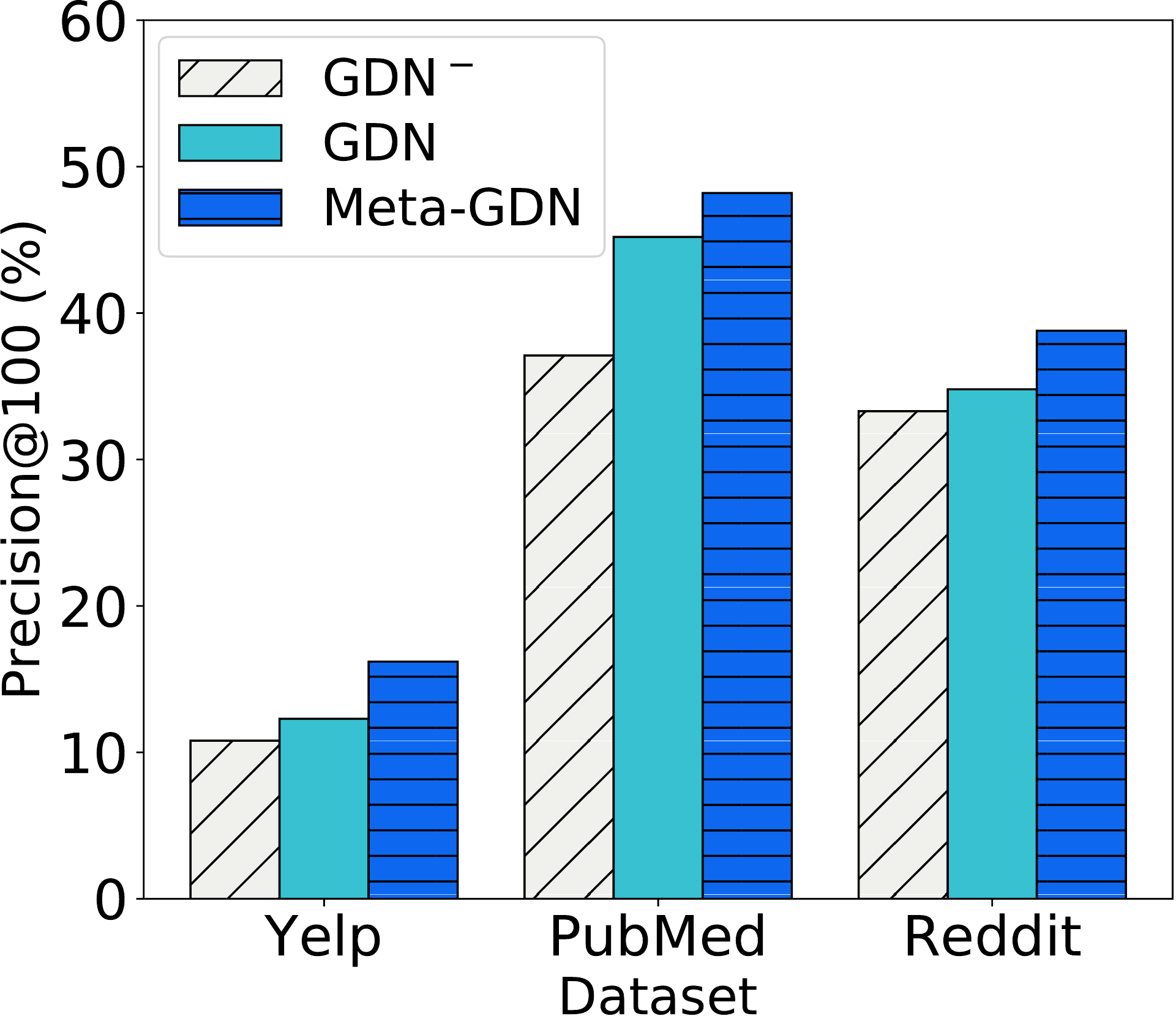}\label{fig:ablation_precision}
    }
    \caption{(a) AUC-ROC results of Meta-GDN and its variants; (b) Precision@100 results of Meta-GDN and its variants.}%
    \label{fig:ablation}%
\end{figure}

% \hide{For instance, the AUC-ROC of Meta-GDN reduces by less than $0.03$ as $r_c$ increases from $0$ to $20\%$.}

% \kz{Actually we haven't introduced GDN-by now, but the figure has the results, we'd better either remove it or replace it with Dominant}

\subsection{Ablation Study (RQ3)}\label{subsec:ablation_robust}
% \qh{add one figure of precision@K}

Moreover, we conduct an ablation study to better examine the contribution of each key component in the proposed framework. In addition to Meta-GDN and its base model GDN, we include another variant GDN$^-$ that excludes the network encoder and cross-network meta-learning in Meta-GDN. We present the results of AUC-ROC and Precision@100 in Figure~\ref{fig:ablation_roc} and Figure~\ref{fig:ablation_precision}, respectively. The corresponding observations are two-fold: \textbf{(1)} by incorporating GNN-based network encoder, GDN largely outperforms GDN$^-$ in anomaly detection on the target network. For example, GDN achieves $8.1\%$ performance improvement over GDN$^-$ on \textit{PubMed} in terms of precision@100. The main reason is that the GNN-based network encoder is able to extract topological information of nodes and to learn highly expressive node representations; and \textbf{(2)} the complete framework Meta-GDN performs consistently better than the base model GDN on all the three datasets. For instance, Meta-GDN improves AUC-ROC by $5.75\% \hide{0.039}$ over GDN on \textit{Yelp} dataset, which verifies the effectiveness of the proposed cross-network meta-learning algorithm for extracting and transferring meta-knowledge across multiple auxiliary networks.

\section{Conclusion}

In this paper, we make the first investigation on the problem of few-shot cross-network anomaly detection. To tackle this problem, we first design a novel GNN architecture, GDN, which is capable of leveraging limited labeled anomalies to enforce statistically significant deviations between abnormal and normal nodes on an individual network. To further utilize the knowledge from auxiliary networks and enable few-shot anomaly detection on the target network, we propose a cross-network meta-learning approach, Meta-GDN, which is able to extract comprehensive meta-knowledge from multiple auxiliary networks in the same domain of the target network. Through extensive experimental evaluations, we demonstrate the superiority of Meta-GDN over the state-of-the-art methods. 
\section*{Acknowledgement}
This work is partially supported by NSF (2029044, 1947135 and 1939725) and ONR (N00014-21-1-4002).

% The second and third authors are partially supported by NSF ( %career
%  % FAI). 

% \input{06conclusion.tex}

% \keywords{datasets, neural networks, gaze detection, text tagging}
% \balance
\bibliographystyle{ACM-Reference-Format}
\bibliography{reference}


\begin{thebibliography}{54}

%%% the Web address.

\ifx \showCODEN    \undefined \def \showCODEN     #1{\unskip}     \fi
\ifx \showDOI      \undefined \def \showDOI       #1{#1}\fi
\ifx \showISBNx    \undefined \def \showISBNx     #1{\unskip}     \fi
\ifx \showISBNxiii \undefined \def \showISBNxiii  #1{\unskip}     \fi
\ifx \showISSN     \undefined \def \showISSN      #1{\unskip}     \fi
\ifx \showLCCN     \undefined \def \showLCCN      #1{\unskip}     \fi
\ifx \shownote     \undefined \def \shownote      #1{#1}          \fi
\ifx \showarticletitle \undefined \def \showarticletitle #1{#1}   \fi
\ifx \showURL      \undefined \def \showURL       {\relax}        \fi
% invisible to TeX
\providecommand\bibfield[2]{#2}
\providecommand\bibinfo[2]{#2}
\providecommand\natexlab[1]{#1}
\providecommand\showeprint[2][]{arXiv:#2}

\bibitem[\protect\citeauthoryear{Akoglu, Tong, and Koutra}{Akoglu
  et~al\mbox{.}}{2015}]%
        {akoglu2015graph}
\bibfield{author}{\bibinfo{person}{Leman Akoglu}, \bibinfo{person}{Hanghang
  Tong}, {and} \bibinfo{person}{Danai Koutra}.}
  \bibinfo{year}{2015}\natexlab{}.
\newblock \showarticletitle{Graph based anomaly detection and description: a
  survey}.
\newblock \bibinfo{journal}{\emph{DMKD}} (\bibinfo{year}{2015}).
\newblock


\bibitem[\protect\citeauthoryear{Bandyopadhyay, Lokesh, and
  Murty}{Bandyopadhyay et~al\mbox{.}}{2019}]%
        {bandyopadhyay2019outlier}
\bibfield{author}{\bibinfo{person}{Sambaran Bandyopadhyay}, \bibinfo{person}{N
  Lokesh}, {and} \bibinfo{person}{M~Narasimha Murty}.}
  \bibinfo{year}{2019}\natexlab{}.
\newblock \showarticletitle{Outlier aware network embedding for attributed
  networks}. In \bibinfo{booktitle}{\emph{AAAI}}.
\newblock


\bibitem[\protect\citeauthoryear{Breunig, Kriegel, Ng, and Sander}{Breunig
  et~al\mbox{.}}{2000}]%
        {breunig2000lof}
\bibfield{author}{\bibinfo{person}{Markus~M Breunig},
  \bibinfo{person}{Hans-Peter Kriegel}, \bibinfo{person}{Raymond~T Ng}, {and}
  \bibinfo{person}{J{\"o}rg Sander}.} \bibinfo{year}{2000}\natexlab{}.
\newblock \showarticletitle{LOF: identifying density-based local outliers}. In
  \bibinfo{booktitle}{\emph{SIGMOD}}.
\newblock


\bibitem[\protect\citeauthoryear{Cao, Lu, and Xu}{Cao et~al\mbox{.}}{2016}]%
        {cao2016deep}
\bibfield{author}{\bibinfo{person}{Shaosheng Cao}, \bibinfo{person}{Wei Lu},
  {and} \bibinfo{person}{Qiongkai Xu}.} \bibinfo{year}{2016}\natexlab{}.
\newblock \showarticletitle{Deep neural networks for learning graph
  representations}. In \bibinfo{booktitle}{\emph{AAAI}}.
\newblock


\bibitem[\protect\citeauthoryear{Ding, Li, Agarwal, and Liu}{Ding
  et~al\mbox{.}}{2020a}]%
        {ding2020inductive}
\bibfield{author}{\bibinfo{person}{Kaize Ding}, \bibinfo{person}{Jundong Li},
  \bibinfo{person}{Nitin Agarwal}, {and} \bibinfo{person}{Huan Liu}.}
  \bibinfo{year}{2020}\natexlab{a}.
\newblock \showarticletitle{Inductive anomaly detection on attributed
  networks}. In \bibinfo{booktitle}{\emph{IJCAI}}.
\newblock


\bibitem[\protect\citeauthoryear{Ding, Li, Bhanushali, and Liu}{Ding
  et~al\mbox{.}}{2019b}]%
        {ding2019deep}
\bibfield{author}{\bibinfo{person}{Kaize Ding}, \bibinfo{person}{Jundong Li},
  \bibinfo{person}{Rohit Bhanushali}, {and} \bibinfo{person}{Huan Liu}.}
  \bibinfo{year}{2019}\natexlab{b}.
\newblock \showarticletitle{Deep anomaly detection on attributed networks}. In
  \bibinfo{booktitle}{\emph{SDM}}.
\newblock


\bibitem[\protect\citeauthoryear{Ding, Li, and Liu}{Ding
  et~al\mbox{.}}{2019a}]%
        {ding2019interactive}
\bibfield{author}{\bibinfo{person}{Kaize Ding}, \bibinfo{person}{Jundong Li},
  {and} \bibinfo{person}{Huan Liu}.} \bibinfo{year}{2019}\natexlab{a}.
\newblock \showarticletitle{Interactive anomaly detection on attributed
  networks}. In \bibinfo{booktitle}{\emph{WSDM}}.
\newblock


\bibitem[\protect\citeauthoryear{Ding, Wang, Li, Shu, Liu, and Liu}{Ding
  et~al\mbox{.}}{2020b}]%
        {ding2020graph}
\bibfield{author}{\bibinfo{person}{Kaize Ding}, \bibinfo{person}{Jianling
  Wang}, \bibinfo{person}{Jundong Li}, \bibinfo{person}{Kai Shu},
  \bibinfo{person}{Chenghao Liu}, {and} \bibinfo{person}{Huan Liu}.}
  \bibinfo{year}{2020}\natexlab{b}.
\newblock \showarticletitle{Graph prototypical networks for few-shot learning
  on attributed networks}. In \bibinfo{booktitle}{\emph{CIKM}}.
\newblock


\bibitem[\protect\citeauthoryear{Dou, Liu, Sun, Deng, Peng, and Yu}{Dou
  et~al\mbox{.}}{2020}]%
        {dou2020enhancing}
\bibfield{author}{\bibinfo{person}{Yingtong Dou}, \bibinfo{person}{Zhiwei Liu},
  \bibinfo{person}{Li Sun}, \bibinfo{person}{Yutong Deng}, \bibinfo{person}{Hao
  Peng}, {and} \bibinfo{person}{Philip~S Yu}.} \bibinfo{year}{2020}\natexlab{}.
\newblock \showarticletitle{Enhancing graph neural network-based fraud
  detectors against camouflaged fraudsters}. In
  \bibinfo{booktitle}{\emph{CIKM}}.
\newblock


\bibitem[\protect\citeauthoryear{Finn, Abbeel, and Levine}{Finn
  et~al\mbox{.}}{2017}]%
        {finn2017model}
\bibfield{author}{\bibinfo{person}{Chelsea Finn}, \bibinfo{person}{Pieter
  Abbeel}, {and} \bibinfo{person}{Sergey Levine}.}
  \bibinfo{year}{2017}\natexlab{}.
\newblock \showarticletitle{Model-agnostic meta-learning for fast adaptation of
  deep networks}.
\newblock \bibinfo{journal}{\emph{ICML}} (\bibinfo{year}{2017}).
\newblock


\bibitem[\protect\citeauthoryear{Hadsell, Chopra, and LeCun}{Hadsell
  et~al\mbox{.}}{2006}]%
        {hadsell2006dimensionality}
\bibfield{author}{\bibinfo{person}{Raia Hadsell}, \bibinfo{person}{Sumit
  Chopra}, {and} \bibinfo{person}{Yann LeCun}.}
  \bibinfo{year}{2006}\natexlab{}.
\newblock \showarticletitle{Dimensionality reduction by learning an invariant
  mapping}. In \bibinfo{booktitle}{\emph{CVPR}}.
\newblock


\bibitem[\protect\citeauthoryear{Hamilton, Ying, and Leskovec}{Hamilton
  et~al\mbox{.}}{2017}]%
        {hamilton2017inductive}
\bibfield{author}{\bibinfo{person}{Will Hamilton}, \bibinfo{person}{Zhitao
  Ying}, {and} \bibinfo{person}{Jure Leskovec}.}
  \bibinfo{year}{2017}\natexlab{}.
\newblock \showarticletitle{Inductive representation learning on large graphs}.
  In \bibinfo{booktitle}{\emph{NeurIPS}}.
\newblock


\bibitem[\protect\citeauthoryear{Hochreiter, Younger, and Conwell}{Hochreiter
  et~al\mbox{.}}{2001}]%
        {hochreiter2001learning}
\bibfield{author}{\bibinfo{person}{Sepp Hochreiter}, \bibinfo{person}{A~Steven
  Younger}, {and} \bibinfo{person}{Peter~R Conwell}.}
  \bibinfo{year}{2001}\natexlab{}.
\newblock \showarticletitle{Learning to learn using gradient descent}. In
  \bibinfo{booktitle}{\emph{ICANN}}.
\newblock


\bibitem[\protect\citeauthoryear{Kipf and Welling}{Kipf and Welling}{2017}]%
        {kipf2017semi}
\bibfield{author}{\bibinfo{person}{Thomas~N. Kipf} {and} \bibinfo{person}{Max
  Welling}.} \bibinfo{year}{2017}\natexlab{}.
\newblock \showarticletitle{Semi-Supervised Classification with Graph
  Convolutional Networks}. In \bibinfo{booktitle}{\emph{ICLR}}.
\newblock


\bibitem[\protect\citeauthoryear{Kriegel, Kroger, Schubert, and Zimek}{Kriegel
  et~al\mbox{.}}{2011}]%
        {kriegel2011interpreting}
\bibfield{author}{\bibinfo{person}{Hans-Peter Kriegel}, \bibinfo{person}{Peer
  Kroger}, \bibinfo{person}{Erich Schubert}, {and} \bibinfo{person}{Arthur
  Zimek}.} \bibinfo{year}{2011}\natexlab{}.
\newblock \showarticletitle{Interpreting and unifying outlier scores}. In
  \bibinfo{booktitle}{\emph{SDM}}.
\newblock


\bibitem[\protect\citeauthoryear{Li, Qin, Liu, Yang, and Li}{Li
  et~al\mbox{.}}{2019b}]%
        {li2019spam}
\bibfield{author}{\bibinfo{person}{Ao Li}, \bibinfo{person}{Zhou Qin},
  \bibinfo{person}{Runshi Liu}, \bibinfo{person}{Yiqun Yang}, {and}
  \bibinfo{person}{Dong Li}.} \bibinfo{year}{2019}\natexlab{b}.
\newblock \showarticletitle{Spam review detection with graph convolutional
  networks}. In \bibinfo{booktitle}{\emph{CIKM}}.
\newblock


\bibitem[\protect\citeauthoryear{Li, Dani, Hu, and Liu}{Li
  et~al\mbox{.}}{2017}]%
        {li2017radar}
\bibfield{author}{\bibinfo{person}{Jundong Li}, \bibinfo{person}{Harsh Dani},
  \bibinfo{person}{Xia Hu}, {and} \bibinfo{person}{Huan Liu}.}
  \bibinfo{year}{2017}\natexlab{}.
\newblock \showarticletitle{Radar: Residual Analysis for Anomaly Detection in
  Attributed Networks.}. In \bibinfo{booktitle}{\emph{IJCAI}}.
\newblock


\bibitem[\protect\citeauthoryear{Li, Huang, Li, Du, and Zou}{Li
  et~al\mbox{.}}{2019a}]%
        {li2019specae}
\bibfield{author}{\bibinfo{person}{Yuening Li}, \bibinfo{person}{Xiao Huang},
  \bibinfo{person}{Jundong Li}, \bibinfo{person}{Mengnan Du}, {and}
  \bibinfo{person}{Na Zou}.} \bibinfo{year}{2019}\natexlab{a}.
\newblock \showarticletitle{SpecAE: Spectral AutoEncoder for Anomaly Detection
  in Attributed Networks}. In \bibinfo{booktitle}{\emph{CIKM}}.
\newblock


\bibitem[\protect\citeauthoryear{Liu, Zhou, Long, Jiang, Dong, and Zhang}{Liu
  et~al\mbox{.}}{2021}]%
        {liu2021isometric}
\bibfield{author}{\bibinfo{person}{Lu Liu}, \bibinfo{person}{Tianyi Zhou},
  \bibinfo{person}{Guodong Long}, \bibinfo{person}{Jing Jiang},
  \bibinfo{person}{Xuanyi Dong}, {and} \bibinfo{person}{Chengqi Zhang}.}
  \bibinfo{year}{2021}\natexlab{}.
\newblock \showarticletitle{Isometric Propagation Network for Generalized
  Zero-shot Learning}. In \bibinfo{booktitle}{\emph{ICLR}}.
\newblock


\bibitem[\protect\citeauthoryear{Liu, Zhou, Long, Jiang, Yao, and Zhang}{Liu
  et~al\mbox{.}}{2019}]%
        {liu2019prototype}
\bibfield{author}{\bibinfo{person}{Lu Liu}, \bibinfo{person}{Tianyi Zhou},
  \bibinfo{person}{Guodong Long}, \bibinfo{person}{Jing Jiang},
  \bibinfo{person}{Lina Yao}, {and} \bibinfo{person}{Chengqi Zhang}.}
  \bibinfo{year}{2019}\natexlab{}.
\newblock \showarticletitle{Prototype propagation networks (PPN) for
  weakly-supervised few-shot learning on category graph}. In
  \bibinfo{booktitle}{\emph{NeurIPS}}.
\newblock


\bibitem[\protect\citeauthoryear{Manning, Sch{\"u}tze, and Raghavan}{Manning
  et~al\mbox{.}}{2008}]%
        {manning2008introduction}
\bibfield{author}{\bibinfo{person}{Christopher~D Manning},
  \bibinfo{person}{Hinrich Sch{\"u}tze}, {and} \bibinfo{person}{Prabhakar
  Raghavan}.} \bibinfo{year}{2008}\natexlab{}.
\newblock \bibinfo{booktitle}{\emph{Introduction to information retrieval}}.
\newblock \bibinfo{publisher}{Cambridge university press}.
\newblock


\bibitem[\protect\citeauthoryear{M{\"u}ller, S{\'a}nchez, M{\"u}lle, and
  B{\"o}hm}{M{\"u}ller et~al\mbox{.}}{2013}]%
        {muller2013ranking}
\bibfield{author}{\bibinfo{person}{Emmanuel M{\"u}ller},
  \bibinfo{person}{Patricia~Iglesias S{\'a}nchez}, \bibinfo{person}{Yvonne
  M{\"u}lle}, {and} \bibinfo{person}{Klemens B{\"o}hm}.}
  \bibinfo{year}{2013}\natexlab{}.
\newblock \showarticletitle{Ranking outlier nodes in subspaces of attributed
  graphs}. In \bibinfo{booktitle}{\emph{ICDE Workshop}}.
\newblock


\bibitem[\protect\citeauthoryear{Pang, Shen, and van~den Hengel}{Pang
  et~al\mbox{.}}{2019}]%
        {pang2019deep}
\bibfield{author}{\bibinfo{person}{Guansong Pang}, \bibinfo{person}{Chunhua
  Shen}, {and} \bibinfo{person}{Anton van~den Hengel}.}
  \bibinfo{year}{2019}\natexlab{}.
\newblock \showarticletitle{Deep anomaly detection with deviation networks}. In
  \bibinfo{booktitle}{\emph{KDD}}.
\newblock


\bibitem[\protect\citeauthoryear{Rayana and Akoglu}{Rayana and Akoglu}{2015}]%
        {rayana2015collective}
\bibfield{author}{\bibinfo{person}{Shebuti Rayana} {and} \bibinfo{person}{Leman
  Akoglu}.} \bibinfo{year}{2015}\natexlab{}.
\newblock \showarticletitle{Collective opinion spam detection: Bridging review
  networks and metadata}. In \bibinfo{booktitle}{\emph{KDD}}.
\newblock


\bibitem[\protect\citeauthoryear{Ruff, Vandermeulen, G{\"o}rnitz, Binder,
  M{\"u}ller, M{\"u}ller, and Kloft}{Ruff et~al\mbox{.}}{2020}]%
        {ruff2019deep}
\bibfield{author}{\bibinfo{person}{Lukas Ruff}, \bibinfo{person}{Robert~A
  Vandermeulen}, \bibinfo{person}{Nico G{\"o}rnitz}, \bibinfo{person}{Alexander
  Binder}, \bibinfo{person}{Emmanuel M{\"u}ller}, \bibinfo{person}{Klaus-Robert
  M{\"u}ller}, {and} \bibinfo{person}{Marius Kloft}.}
  \bibinfo{year}{2020}\natexlab{}.
\newblock \showarticletitle{Deep Semi-Supervised Anomaly Detection}. In
  \bibinfo{booktitle}{\emph{ICLR}}.
\newblock


\bibitem[\protect\citeauthoryear{S{\'a}nchez, M{\"u}ller, Irmler, and
  B{\"o}hm}{S{\'a}nchez et~al\mbox{.}}{2014}]%
        {sanchez2014local}
\bibfield{author}{\bibinfo{person}{Patricia~Iglesias S{\'a}nchez},
  \bibinfo{person}{Emmanuel M{\"u}ller}, \bibinfo{person}{Oretta Irmler}, {and}
  \bibinfo{person}{Klemens B{\"o}hm}.} \bibinfo{year}{2014}\natexlab{}.
\newblock \showarticletitle{Local context selection for outlier ranking in
  graphs with multiple numeric node attributes}. In
  \bibinfo{booktitle}{\emph{SSDBM}}.
\newblock


\bibitem[\protect\citeauthoryear{Santoro, Bartunov, Botvinick, Wierstra, and
  Lillicrap}{Santoro et~al\mbox{.}}{2016}]%
        {santoro2016meta}
\bibfield{author}{\bibinfo{person}{Adam Santoro}, \bibinfo{person}{Sergey
  Bartunov}, \bibinfo{person}{Matthew Botvinick}, \bibinfo{person}{Daan
  Wierstra}, {and} \bibinfo{person}{Timothy Lillicrap}.}
  \bibinfo{year}{2016}\natexlab{}.
\newblock \showarticletitle{Meta-learning with memory-augmented neural
  networks}. In \bibinfo{booktitle}{\emph{ICML}}.
\newblock


\bibitem[\protect\citeauthoryear{Sen, Namata, Bilgic, Getoor, Galligher, and
  Eliassi-Rad}{Sen et~al\mbox{.}}{2008}]%
        {sen2008collective}
\bibfield{author}{\bibinfo{person}{Prithviraj Sen}, \bibinfo{person}{Galileo
  Namata}, \bibinfo{person}{Mustafa Bilgic}, \bibinfo{person}{Lise Getoor},
  \bibinfo{person}{Brian Galligher}, {and} \bibinfo{person}{Tina Eliassi-Rad}.}
  \bibinfo{year}{2008}\natexlab{}.
\newblock \showarticletitle{Collective classification in network data}.
\newblock \bibinfo{journal}{\emph{AI magazine}} (\bibinfo{year}{2008}).
\newblock


\bibitem[\protect\citeauthoryear{Shen, Dai, Chung, Lu, and Choi}{Shen
  et~al\mbox{.}}{2020}]%
        {shen2020adversarial}
\bibfield{author}{\bibinfo{person}{Xiao Shen}, \bibinfo{person}{Quanyu Dai},
  \bibinfo{person}{Fu-lai Chung}, \bibinfo{person}{Wei Lu}, {and}
  \bibinfo{person}{Kup-Sze Choi}.} \bibinfo{year}{2020}\natexlab{}.
\newblock \showarticletitle{Adversarial Deep Network Embedding for
  Cross-Network Node Classification.}. In \bibinfo{booktitle}{\emph{AAAI}}.
\newblock


\bibitem[\protect\citeauthoryear{Skillicorn}{Skillicorn}{2007}]%
        {skillicorn2007detecting}
\bibfield{author}{\bibinfo{person}{David~B Skillicorn}.}
  \bibinfo{year}{2007}\natexlab{}.
\newblock \showarticletitle{Detecting anomalies in graphs}. In
  \bibinfo{booktitle}{\emph{ISI}}.
\newblock


\bibitem[\protect\citeauthoryear{Song, Wu, Jermaine, and Ranka}{Song
  et~al\mbox{.}}{2007}]%
        {song2007conditional}
\bibfield{author}{\bibinfo{person}{Xiuyao Song}, \bibinfo{person}{Mingxi Wu},
  \bibinfo{person}{Christopher Jermaine}, {and} \bibinfo{person}{Sanjay
  Ranka}.} \bibinfo{year}{2007}\natexlab{}.
\newblock \showarticletitle{Conditional anomaly detection}.
\newblock \bibinfo{journal}{\emph{TKDE}} (\bibinfo{year}{2007}).
\newblock


\bibitem[\protect\citeauthoryear{Tang, Zhang, Yao, Li, Zhang, and Su}{Tang
  et~al\mbox{.}}{2008}]%
        {tang2008arnetminer}
\bibfield{author}{\bibinfo{person}{Jie Tang}, \bibinfo{person}{Jing Zhang},
  \bibinfo{person}{Limin Yao}, \bibinfo{person}{Juanzi Li}, \bibinfo{person}{Li
  Zhang}, {and} \bibinfo{person}{Zhong Su}.} \bibinfo{year}{2008}\natexlab{}.
\newblock \showarticletitle{Arnetminer: extraction and mining of academic
  social networks}. In \bibinfo{booktitle}{\emph{KDD}}.
\newblock


\bibitem[\protect\citeauthoryear{Tang, Li, Sun, Yao, Mitra, and Wang}{Tang
  et~al\mbox{.}}{2020}]%
        {tang2019transferring}
\bibfield{author}{\bibinfo{person}{Xianfeng Tang}, \bibinfo{person}{Yandong
  Li}, \bibinfo{person}{Yiwei Sun}, \bibinfo{person}{Huaxiu Yao},
  \bibinfo{person}{Prasenjit Mitra}, {and} \bibinfo{person}{Suhang Wang}.}
  \bibinfo{year}{2020}\natexlab{}.
\newblock \showarticletitle{Transferring Robustness for Graph Neural Network
  Against Poisoning Attacks}. In \bibinfo{booktitle}{\emph{WSDM}}.
\newblock


\bibitem[\protect\citeauthoryear{Tong and Lin}{Tong and Lin}{2011}]%
        {tong2011non}
\bibfield{author}{\bibinfo{person}{Hanghang Tong} {and}
  \bibinfo{person}{Ching-Yung Lin}.} \bibinfo{year}{2011}\natexlab{}.
\newblock \showarticletitle{Non-negative residual matrix factorization with
  application to graph anomaly detection}. In \bibinfo{booktitle}{\emph{SDM}}.
\newblock


\bibitem[\protect\citeauthoryear{Veli{\v{c}}kovi{\'c}, Cucurull, Casanova,
  Romero, Lio, and Bengio}{Veli{\v{c}}kovi{\'c} et~al\mbox{.}}{2018}]%
        {velickovic2017graph}
\bibfield{author}{\bibinfo{person}{Petar Veli{\v{c}}kovi{\'c}},
  \bibinfo{person}{Guillem Cucurull}, \bibinfo{person}{Arantxa Casanova},
  \bibinfo{person}{Adriana Romero}, \bibinfo{person}{Pietro Lio}, {and}
  \bibinfo{person}{Yoshua Bengio}.} \bibinfo{year}{2018}\natexlab{}.
\newblock \showarticletitle{Graph attention networks}. In
  \bibinfo{booktitle}{\emph{ICLR}}.
\newblock


\bibitem[\protect\citeauthoryear{Vinyals, Blundell, Lillicrap, Wierstra,
  et~al\mbox{.}}{Vinyals et~al\mbox{.}}{2016}]%
        {vinyals2016matching}
\bibfield{author}{\bibinfo{person}{Oriol Vinyals}, \bibinfo{person}{Charles
  Blundell}, \bibinfo{person}{Timothy Lillicrap}, \bibinfo{person}{Daan
  Wierstra}, {et~al\mbox{.}}} \bibinfo{year}{2016}\natexlab{}.
\newblock \showarticletitle{Matching networks for one shot learning}. In
  \bibinfo{booktitle}{\emph{NeurIPS}}.
\newblock


\bibitem[\protect\citeauthoryear{Wang, Lin, Cui, Jia, Wang, Fang, Yu, Zhou,
  Yang, and Qi}{Wang et~al\mbox{.}}{2019}]%
        {wang2019semi}
\bibfield{author}{\bibinfo{person}{Daixin Wang}, \bibinfo{person}{Jianbin Lin},
  \bibinfo{person}{Peng Cui}, \bibinfo{person}{Quanhui Jia},
  \bibinfo{person}{Zhen Wang}, \bibinfo{person}{Yanming Fang},
  \bibinfo{person}{Quan Yu}, \bibinfo{person}{Jun Zhou},
  \bibinfo{person}{Shuang Yang}, {and} \bibinfo{person}{Yuan Qi}.}
  \bibinfo{year}{2019}\natexlab{}.
\newblock \showarticletitle{A Semi-supervised Graph Attentive Network for
  Financial Fraud Detection}. In \bibinfo{booktitle}{\emph{ICDM}}.
\newblock


\bibitem[\protect\citeauthoryear{Wang, Luo, Ding, Zhang, Li, and Zheng}{Wang
  et~al\mbox{.}}{2020}]%
        {wang2020graph}
\bibfield{author}{\bibinfo{person}{Ning Wang}, \bibinfo{person}{Minnan Luo},
  \bibinfo{person}{Kaize Ding}, \bibinfo{person}{Lingling Zhang},
  \bibinfo{person}{Jundong Li}, {and} \bibinfo{person}{Qinghua Zheng}.}
  \bibinfo{year}{2020}\natexlab{}.
\newblock \showarticletitle{Graph Few-shot Learning with Attribute Matching}.
  In \bibinfo{booktitle}{\emph{CIKM}}.
\newblock


\bibitem[\protect\citeauthoryear{Wu, Zhang, Souza~Jr, Fifty, Yu, and
  Weinberger}{Wu et~al\mbox{.}}{2019b}]%
        {wu2019simplifying}
\bibfield{author}{\bibinfo{person}{Felix Wu}, \bibinfo{person}{Tianyi Zhang},
  \bibinfo{person}{Amauri Holanda~de Souza~Jr}, \bibinfo{person}{Christopher
  Fifty}, \bibinfo{person}{Tao Yu}, {and} \bibinfo{person}{Kilian~Q
  Weinberger}.} \bibinfo{year}{2019}\natexlab{b}.
\newblock \showarticletitle{Simplifying graph convolutional networks}. In
  \bibinfo{booktitle}{\emph{ICML}}.
\newblock


\bibitem[\protect\citeauthoryear{Wu, Pan, Du, Tsang, Zhu, and Du}{Wu
  et~al\mbox{.}}{2019a}]%
        {wu2019long}
\bibfield{author}{\bibinfo{person}{Man Wu}, \bibinfo{person}{Shirui Pan},
  \bibinfo{person}{Lan Du}, \bibinfo{person}{Ivor Tsang},
  \bibinfo{person}{Xingquan Zhu}, {and} \bibinfo{person}{Bo Du}.}
  \bibinfo{year}{2019}\natexlab{a}.
\newblock \showarticletitle{Long-short Distance Aggregation Networks for
  Positive Unlabeled Graph Learning}. In \bibinfo{booktitle}{\emph{CIKM}}.
\newblock


\bibitem[\protect\citeauthoryear{Wu, Pan, Zhou, Chang, and Zhu}{Wu
  et~al\mbox{.}}{2020}]%
        {wu2020unsupervised}
\bibfield{author}{\bibinfo{person}{Man Wu}, \bibinfo{person}{Shirui Pan},
  \bibinfo{person}{Chuan Zhou}, \bibinfo{person}{Xiaojun Chang}, {and}
  \bibinfo{person}{Xingquan Zhu}.} \bibinfo{year}{2020}\natexlab{}.
\newblock \showarticletitle{Unsupervised Domain Adaptive Graph Convolutional
  Networks}. In \bibinfo{booktitle}{\emph{The Web Conference}}.
\newblock


\bibitem[\protect\citeauthoryear{Xu, Hu, Leskovec, and Jegelka}{Xu
  et~al\mbox{.}}{2019}]%
        {xu2018powerful}
\bibfield{author}{\bibinfo{person}{Keyulu Xu}, \bibinfo{person}{Weihua Hu},
  \bibinfo{person}{Jure Leskovec}, {and} \bibinfo{person}{Stefanie Jegelka}.}
  \bibinfo{year}{2019}\natexlab{}.
\newblock \showarticletitle{How powerful are graph neural networks?}. In
  \bibinfo{booktitle}{\emph{ICLR}}.
\newblock


\bibitem[\protect\citeauthoryear{Xu, Yuruk, Feng, and Schweiger}{Xu
  et~al\mbox{.}}{2007}]%
        {xu2007scan}
\bibfield{author}{\bibinfo{person}{Xiaowei Xu}, \bibinfo{person}{Nurcan Yuruk},
  \bibinfo{person}{Zhidan Feng}, {and} \bibinfo{person}{Thomas~AJ Schweiger}.}
  \bibinfo{year}{2007}\natexlab{}.
\newblock \showarticletitle{Scan: a structural clustering algorithm for
  networks}. In \bibinfo{booktitle}{\emph{Proceedings of the 13th ACM SIGKDD
  International Conference on Knowledge Discovery and Data mining (KDD)}}.
\newblock


\bibitem[\protect\citeauthoryear{You, Liu, Ying, Pande, and Leskovec}{You
  et~al\mbox{.}}{2018}]%
        {you2018graph}
\bibfield{author}{\bibinfo{person}{Jiaxuan You}, \bibinfo{person}{Bowen Liu},
  \bibinfo{person}{Zhitao Ying}, \bibinfo{person}{Vijay Pande}, {and}
  \bibinfo{person}{Jure Leskovec}.} \bibinfo{year}{2018}\natexlab{}.
\newblock \showarticletitle{Graph convolutional policy network for
  goal-directed molecular graph generation}. In
  \bibinfo{booktitle}{\emph{NeurIPS}}.
\newblock


\bibitem[\protect\citeauthoryear{Zafarani, Abbasi, and Liu}{Zafarani
  et~al\mbox{.}}{2014}]%
        {zafarani2014social}
\bibfield{author}{\bibinfo{person}{Reza Zafarani},
  \bibinfo{person}{Mohammad~Ali Abbasi}, {and} \bibinfo{person}{Huan Liu}.}
  \bibinfo{year}{2014}\natexlab{}.
\newblock \bibinfo{booktitle}{\emph{Social media mining: an introduction}}.
\newblock \bibinfo{publisher}{Cambridge University Press}.
\newblock


\bibitem[\protect\citeauthoryear{Zhang, Jin, and Zhou}{Zhang
  et~al\mbox{.}}{2010}]%
        {zhang2010understanding}
\bibfield{author}{\bibinfo{person}{Yin Zhang}, \bibinfo{person}{Rong Jin},
  {and} \bibinfo{person}{Zhi-Hua Zhou}.} \bibinfo{year}{2010}\natexlab{}.
\newblock \showarticletitle{Understanding bag-of-words model: a statistical
  framework}.
\newblock \bibinfo{journal}{\emph{IJMLC}} (\bibinfo{year}{2010}).
\newblock


\bibitem[\protect\citeauthoryear{Zhang, Zhao, Jiang, and Zhou}{Zhang
  et~al\mbox{.}}{2019}]%
        {zhang2019learning}
\bibfield{author}{\bibinfo{person}{Zhen-Yu Zhang}, \bibinfo{person}{Peng Zhao},
  \bibinfo{person}{Yuan Jiang}, {and} \bibinfo{person}{Zhi-Hua Zhou}.}
  \bibinfo{year}{2019}\natexlab{}.
\newblock \showarticletitle{Learning from incomplete and inaccurate
  supervision}. In \bibinfo{booktitle}{\emph{KDD}}.
\newblock


\bibitem[\protect\citeauthoryear{Zhao, Deng, Yu, Jiang, Wang, and Jiang}{Zhao
  et~al\mbox{.}}{2020}]%
        {zhao2020error}
\bibfield{author}{\bibinfo{person}{Tong Zhao}, \bibinfo{person}{Chuchen Deng},
  \bibinfo{person}{Kaifeng Yu}, \bibinfo{person}{Tianwen Jiang},
  \bibinfo{person}{Daheng Wang}, {and} \bibinfo{person}{Meng Jiang}.}
  \bibinfo{year}{2020}\natexlab{}.
\newblock \showarticletitle{Error-Bounded Graph Anomaly Loss for GNNs}. In
  \bibinfo{booktitle}{\emph{CIKM}}.
\newblock


\bibitem[\protect\citeauthoryear{Zhou and Paffenroth}{Zhou and
  Paffenroth}{2017}]%
        {zhou2017anomaly}
\bibfield{author}{\bibinfo{person}{Chong Zhou} {and} \bibinfo{person}{Randy~C
  Paffenroth}.} \bibinfo{year}{2017}\natexlab{}.
\newblock \showarticletitle{Anomaly detection with robust deep autoencoders}.
  In \bibinfo{booktitle}{\emph{KDD}}.
\newblock


\bibitem[\protect\citeauthoryear{Zhou, He, Yang, and Fan}{Zhou
  et~al\mbox{.}}{2018}]%
        {zhou2018sparc}
\bibfield{author}{\bibinfo{person}{Dawei Zhou}, \bibinfo{person}{Jingrui He},
  \bibinfo{person}{Hongxia Yang}, {and} \bibinfo{person}{Wei Fan}.}
  \bibinfo{year}{2018}\natexlab{}.
\newblock \showarticletitle{Sparc: Self-paced network representation for
  few-shot rare category characterization}. In \bibinfo{booktitle}{\emph{KDD}}.
\newblock


\bibitem[\protect\citeauthoryear{Zhou, Cao, Trajcevski, Zhang, Zhong, and
  Geng}{Zhou et~al\mbox{.}}{2020}]%
        {zhou2020fast}
\bibfield{author}{\bibinfo{person}{Fan Zhou}, \bibinfo{person}{Chengtai Cao},
  \bibinfo{person}{Goce Trajcevski}, \bibinfo{person}{Kunpeng Zhang},
  \bibinfo{person}{Ting Zhong}, {and} \bibinfo{person}{Ji Geng}.}
  \bibinfo{year}{2020}\natexlab{}.
\newblock \showarticletitle{Fast Network Alignment via Graph Meta-Learning}. In
  \bibinfo{booktitle}{\emph{INFOCOM}}.
\newblock


\bibitem[\protect\citeauthoryear{Zhou, Li, Cao, Ying, and Tong}{Zhou
  et~al\mbox{.}}{2019b}]%
        {zhou2019admiring}
\bibfield{author}{\bibinfo{person}{Qinghai Zhou}, \bibinfo{person}{Liangyue
  Li}, \bibinfo{person}{Nan Cao}, \bibinfo{person}{Lei Ying}, {and}
  \bibinfo{person}{Hanghang Tong}.} \bibinfo{year}{2019}\natexlab{b}.
\newblock \showarticletitle{ADMIRING: Adversarial multi-network mining}. In
  \bibinfo{booktitle}{\emph{ICDM}}.
\newblock


\bibitem[\protect\citeauthoryear{Zhou, Li, and Tong}{Zhou
  et~al\mbox{.}}{2019a}]%
        {zhou2019towards}
\bibfield{author}{\bibinfo{person}{Qinghai Zhou}, \bibinfo{person}{Liangyue
  Li}, {and} \bibinfo{person}{Hanghang Tong}.}
  \bibinfo{year}{2019}\natexlab{a}.
\newblock \showarticletitle{Towards Real Time Team Optimization}. In
  \bibinfo{booktitle}{\emph{Big Data}}.
\newblock


\bibitem[\protect\citeauthoryear{Z{\"u}gner and G{\"u}nnemann}{Z{\"u}gner and
  G{\"u}nnemann}{2019}]%
        {zugner2019adversarial}
\bibfield{author}{\bibinfo{person}{Daniel Z{\"u}gner} {and}
  \bibinfo{person}{Stephan G{\"u}nnemann}.} \bibinfo{year}{2019}\natexlab{}.
\newblock \showarticletitle{Adversarial Attacks on Graph Neural Networks via
  Meta Learning}. In \bibinfo{booktitle}{\emph{ICLR}}.
\newblock


\end{thebibliography}

\vfill\eject
\newpage

\end{document}